\documentclass[journal]{IEEEtran}

\usepackage{cite}
\usepackage[numbers,sort&compress]{natbib}
\usepackage{amsmath,amssymb,amsfonts, bm}
\usepackage{algorithm}
\usepackage{algorithmic}
\usepackage{subfigure}
\usepackage{graphicx}
\usepackage{caption}
\usepackage{textcomp}
\usepackage{verbatim}
\usepackage{xcolor}
\usepackage{authblk}
\usepackage[marginal]{footmisc}
\usepackage{multirow}
\usepackage{multicol}
\def\BibTeX{{\rm B\kern-.05em{\sc i\kern-.025em b}\kern-.08em
    T\kern-.1667em\lower.7ex\hbox{E}\kern-.125emX}}

\title{The Similarity-Consensus Regularized Multi-view Learning for Dimension Reduction}
\date{July 2019}
\author[1]{Xiangzhu Meng}
\author[2]{Huibing Wang}
\author[1]{Lin Feng}
\affil[1]{School of Computer Science and Technology, Dalian University of Technology, Dalian, China \authorcr fenglin@dlut.edu.cn, xiangzhu\_meng@mail.dlut.edu.cn}
\affil[2]{Information Science and Technology College, Dalian Maritime University, Dalian, China \authorcr huibing.wang@dlmu.edu.cn}

\begin{document}

\maketitle
\begin{abstract}
During the last decades, learning a low-dimensional space with discriminative information for dimension reduction (DR) has gained a surge of interest. However, it's not accessible for these DR methods to achieve satisfactory performance when facing the features from multiple views. In multi-view learning problems, one instance can be represented by multiple heterogeneous features, which are highly related but sometimes look different from each other. In addition, correlations between features from multiple views always vary greatly, which challenges the capability of multi-view learning methods. Consequently, constructing a multi-view learning framework with generalization and scalability, which could take advantage of multi-view information as much as possible, is extremely necessary but challenging. To implement the above target, this paper proposes a novel multi-view learning framework based on similarity consensus, which makes full use of correlations among multi-view features while considering the scalability and robustness of the framework. It aims to straightforwardly extend those existing DR methods into multi-view learning domain by preserving the similarity between different views to capture the low-dimensional embedding. Two schemes based on pairwise-consensus and centroid-consensus are separately proposed to force multiple views to learn from each other and then an iterative alternating strategy is developed to obtain the optimal solution. The proposed method is evaluated on 5 benchmark datasets and comprehensive experiments show that our proposed multi-view framework can yield comparable and promising performance with previous approaches proposed in recent literatures.

\end{abstract}


\section{Introduction}\label{introduction}
The raw data are often collected from different kinds of viewpoints \cite{li2018survey, lahat2015multimodal, sun2013survey, xu2013survey} in many real-world applications, such as image retrieval \cite{smeulders2000content,datta2008image}, text categorization \cite{jiang2011fuzzy} and face recognition \cite{tao2012discriminative,qiao2010sparsity}. For example, web pages usually consist of title, page-text and hyperlink information; an image could be described with color, text or shape information, such as HSV, Local Binary Pattern (LBP) \cite{ojala2002multiresolution}, Gist \cite{douze2009evaluation}, Histogram of Gradients (HoG)\cite{dalal2005histograms}, Edge Direction Histogram (EDH) \cite{gao2008image}. Different from single view data which only contains partial information, multi-view data usually carries complementary information among different views. Even though multi-view features usually contain more useful information than single view scenario, high dimensional problems and integration among different views influence the efficiency and performance of the application system. To address the above-mentioned issues, most DR methods and multi-view learning methods are proposed.

To tackle time consuming and computational cost due to high dimensional features, a variety of DR methods are proposed to find a low dimensional space by preserving come properties of raw features. Existing DR methods could be mainly divided into three categories: subspace learning \cite{fukunaga2013introduction,wold1987principal, yu2017robust,dudoit2002comparison, he2004locality, he2005neighborhood, cai2007locality, weinberger2006distance, xu2007marginal, qiao2010sparsity}, kernel learning \cite{scholkopf1997kernel, mika1999fisher,torresani2007large} and manifold learning \cite{tenenbaum2000global,belkin2003laplacian, roweis2000nonlinear, zhang2004principal}. Principal Components Analysis(PCA) \cite{wold1987principal} and Linear Discriminant Analysis(LDA) \cite{dudoit2002comparison,yu2017robust} are two popular subspace learning methods based on linear transform, which maximize the global variance of low dimensional features and the ratio between between-class scatter and within-class scatter. Contrast to PCA and LDA that only maintain global structure, some DR methods aim to find an optimal subspace while could preserve the the local relations between different samples by different means, such as Locality Preserving Projection(LPP) \cite{he2004locality}, Neighborhood Preserving Embedding(NPE) \cite{he2005neighborhood}, Locality Semantic Discriminant Analysis(LSDA) \cite{cai2007locality}, Large Margin Nearest Neighbor(LMNN) \cite{weinberger2006distance}, Margin Fisher Analysis(MFA) \cite{xu2007marginal}, and Sparsity Preserving Projections(SPP) \cite{qiao2010sparsity}. Unlike these linear methods above, kernel methods aim to find a low dimensional space in nonlinearly high-dimensional space by using kernel tricks. There are some represented works including Kernel Principal Components Analysis(KPCA) \cite{scholkopf1997kernel}, Kernel Fisher Discriminant Analysis(KFDA) \cite{mika1999fisher}, and Kernel Large Margin  Component Analysis(KLMCA) \cite{torresani2007large}, which extend PCA, LDA and LMNN into nonlinear subspace learning domain respectively. Except for kernel methods above, manifold learning is an effective approach for nonlinear dimension reduction, which learns an embedded low-dimensional manifold through preserving local geometric information of the original high dimensional space. Representative manifold learning algorithms include Isometric Mapping (Isomap) \cite{tenenbaum2000global}, Laplacian Embedding (LE) \cite{belkin2003laplacian}, Local Linear Embedding (LLE) \cite{roweis2000nonlinear}, and Local Tangent Space Alignment(LSTA)\cite{zhang2004principal}. These DR methods above mainly focus on single view, and couldn't be directly extended to process multi-view cases.

On integrating rich information among different viewpoints, a variety of multi-view learning methods \cite{chaudhuri2009multi, rupnik2010multi, kan2016multi, zhang2018generalized, xia2010multiview, kumar2011co} has been proposed in the past decade. The work \cite{chaudhuri2009multi} proposes that Canonical Correlation Analysis (CCA) \cite{hardoon2004canonical} could be used to project the two view into the common subspace by maximizing the cross correlation between two views. Furthermore, CCA is further generalized for multi-view scenario termed as multi-view canonical correlation analysis (MCCA) \cite{rupnik2010multi}. Multi-View Discriminant Analysis \cite{kan2016multi} is proposed to extend LDA into a multi-view setting, which projects multi-view features to one discriminative common subspace. The paper \cite{zhang2018generalized} proposes a Generalized Latent Multi-View Subspace Clustering, which jointly learns the latent representation and multi-view subspace representation within the unified framework. Besides these multi-view learning methods, some researches based on multiple graph learning have been developed. Multiview Spectral Embedding (MSE) \cite{xia2010multiview} incorporates conventional algorithms with multi-view data to find a common low-dimensional subspace, which exploits low-dimensional representations based on graph. The work \cite{kumar2011co} aims to propose a co-regularized multi-view spectral clustering framework that captures complementary information among different viewpoints by co-regularizing a clustering hypothesis. In addition to these works above, such works in \cite{wang2015robust, wang2016iterative,wang2018multiview, wang2017unsupervised, wu2018deep, wang2017effective,nie2017auto,wu2019cycle,wang2018beyond} also obtain promising performance in multi-view learning environment. Even though these methods have made good progress in integrating multi-view information, limitations of generalization and scalability exist all the time.

\subsection{Motivation}
For many real-world applications, the same object usually could be described at different views, which arises high dimensional problems and integration problems among different views influence. As a famous family of high dimensional data processing, DR methods have attracted wide attention due to their excellent performance. As aspect of integration among different views, some multi-view methods are proposed to achieve considerable performance but some limitations including generalization and scalability still remain. It's very difficult for these work to extent DR methods based on single view with their opinion into multi-view setting, such that we couldn't make the most of advantages of single view based DR methods when integrating rich information among different views.

Based on the discussion above, we decide to comprehensively investigate the multi-view learning problem from the following 2 aspects:
\begin{itemize}
    \item[1] Is it feasible to extend these DR methods, including subspace learning, kernel learning, and manifold learning, to process multi-view problems together?
    \item[2] How to integrate multi-view features from different features under different views?
\end{itemize}

\subsection{Contributions}
In this paper, we first propose a novel multi-view learning framework based on similarity consensus for manifold learning methods, which maintain local linear structure in the geometric manifold space. Especially, we utilize the correlation between similarity matrices of two views as the consensus term and then the pairwise consensus-based framework and the centroid consensus-based framework are designed to integrate different information from multiple views respectively. Then, an optimization algorithm using iterative alternating strategy is developed to obtain the optimal solution of the framework. Furthermore, we extend the framework for subspace learning and kernel learning so that most DR methods based on single view could be extended to achieve the dimension reduction in multi-view features. Finally, the experiments are conducted on 5 benchmark datasets. To sum up, the contributions in this paper are illustrated as follows:
\begin{itemize}
    \item The novel multi-view manifold learning framework based on similarity consensus is proposed to integrate different information from multiple views and then we propose an effective and robust iterative method to seek an optimal solution for the multi-view framework.
    \item We extend single view based subspace learning and kernel learning methods that could be cast as a special form of the quadratically constrained quadratic program into this multi-view framework.
    \item  The experimental results on 5 benchmark datasets demonstrate that the proposed method outperforms its counterparts including approaches proposed in recent literatures and achieve comparable and promising performance.
\end{itemize}

\subsection{Organization}
The rest of the paper is organized as follows. In Section 2, we provide related methods which have attracted extensive attention. In Section 3, we describe the construction procedure of the multi-view learning framework based on similarity consensus for manifold learning and illustrate the optimization algorithm in detail. In Section 4, we extend subspace learning and kernel learning methods based on similarity consensus into the multi-view framework. In Section 5, empirical evaluations based on the applications of text classification and image classification demonstrate the effectiveness of our proposed approach. In Section 6, we make a conclusion of this paper.

\begin{figure*}[htbp]
\centering
\includegraphics[width=0.9\textwidth]{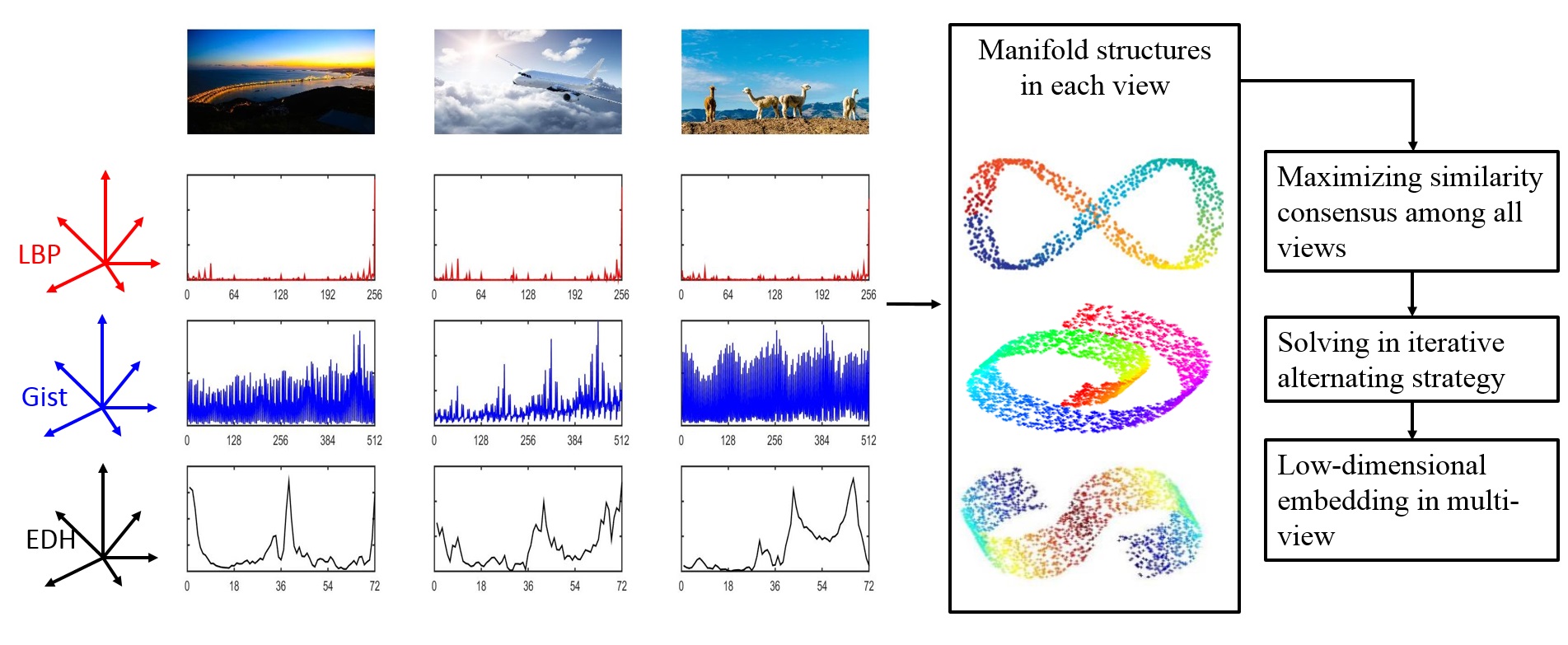}
\caption{The working procedure of Multi-view manifold learning framework}
\label{test}
\end{figure*}

\section{Related work}
In this section, we review two widely concerned multi-view learning methods, including Canonical Correlation Analysis (CCA) \cite{hardoon2004canonical}and Co-regularized multi-view spectral clustering \cite{kumar2011co}, which have attracted wide attention in practical applications.

\subsection{Canonical Correlation Analysis}
Canonical Correlation Analysis (CCA) \cite{hardoon2004canonical} can be seen as the problem of finding basis vectors for two sets of variables such that the correlation between the projections of the variables on to these basis vectors are mutually maximised. Consider the multivariate vectors of the form $(\bm{X}, \bm{Y})$, where $\bm{X}=\left\{ {\bm{x}_1,\bm{x}_{_2}, \ldots ,\bm{x}_N} \right\}$ denotes the samples from one view and $\bm{Y}=\left\{ {\bm{y}_1,\bm{y}_{_2}, \ldots ,\bm{y}_N} \right\}$ denotes the samples from the other view. CCA is to choose to maximize the correlation between the two vectors by projection transform. So we could get the following optimization problem:
\begin{equation}
\label{CCA}
\begin{array}{l}
\mathop {{\rm{max}}}\limits_{{\bm{w}_x},{\bm{w}_y}} {\rm{ }} tr\left( {\bm{w}_x^T\bm{X}\bm{Y}{\bm{w}_y}} \right)\\
s.t. \hspace{1em} \bm{w}_x^TXX{\bm{w}_x} = \bm{I},\hspace{0.5em} \bm{w}_y^T\bm{Y}\bm{Y}{\bm{w}_y} = \bm{I}.
\end{array}
\end{equation}
where $\bm{w}_x$ and $\bm{w}_y$ denote the projecting matrix of the samples $X$ and the samples $Y$ respectively. To solve the above optimization problem, the eigenvalue decomposition method could be employed.

\subsection{Co-regularized Multi-view Spectral Clustering}
Co-regularized Multi-view Spectral Clustering \cite{kumar2011co} is a spectral clustering algorithm under multiple views, which achieves this goal by co-regularizing the clustering hypothesis across views. Assume that given data has multiple views, and let ${\bm{X}^v}=\left\{ {\bm{x}_1^v,\bm{x}_{_{}^2}^v, \ldots ,\bm{x}_N^v} \right\}$ denote the features set in the $v$th view and $\bm{L}^v$ denotes the normalized graph Laplacian matrix in this view. According to the co-regularizing the clustering hypothesis across views, we could get the following  maximization problem for $m$ views:
\begin{equation}\label{co_regularizaton_loss}
\begin{split}
&\mathop {\max }\limits_{{\bm{U}^1},{\bm{U}^2}, \ldots ,{\bm{U}^m} \in {\mathbb{R}^{N \times k}}} \sum\limits_{v = 1}^m {tr({\bm{U}^v}^{^T}{\bm{L}^v}{\bm{U}^v})} \\
& + \gamma \sum\limits_{1 \le v,w \le m} {tr\left( {{\bm{U}^v}{\bm{U}^v}^{^T}{\bm{U}^w}{\bm{U}^w}^{^T}} \right)} \\
&\hspace{3em}s.t.\hspace{1.5em}{\bm{U}^v}^{^T}{\bm{U}^v}{ =  \bm{I},} \forall {\rm{  1}} \le v \le m{\rm{    }}    \\
\end{split}
\end{equation}
where the hyperparameter $\gamma$ trades-off the spectral clustering objectives and the spectral embedding disagreement term, and the second item in the Eq.(\ref{co_regularizaton_loss}) reflects the measure of agreement between two views. To solve the loss objective problem, the iterative alternating strategy could be employed.

\section{The Similarity-Consensus Regularized Multi-view Manifold Learning}
In this section, We first introduce manifold learning methods based on single view and background knowledge towards multi-view manifold learning problems. Then, we utilize the correlation between similarity matrices of two views as the consensus term and propose a multi-view framework based on such consensus terms among multiple views, which not only preserves the strength of DR methods based on single view but also fully utilizes compatible and complementary information from multi-view features in Section\ref{main_framework}. Besides, we also provide a centroid-based scheme to obtain a common low-dimensional embedding. An iterative alternating strategy is adopted to find the optimal solution for our framework solution and the optimization procedure is illustrated in detail in Section \ref{optimization}. And Fig.\ref{test} shows the working procedure of Multi-view manifold learning framework. Finally, we clearly analyze the influences on time complexity to explain the efficiency of our framework.


\subsection{Manifold learning based on single view}
 LLE\cite{roweis2000nonlinear} and LE \cite{belkin2003laplacian} are two representative manifold learning algorithms, which lie on a sub-manifold of the observations space. Inspired by LLE and LE, we focus on the special form of a quadratically constrained quadratic program(QCQP) in this paper. Suppose that we have the $v$th view and $\bm{Y}^v$ denotes the low dimensional embedding to solved. We express $r(\bm{Y}^v \bm{M}^v \bm{Y}^{v^T})$ as objective function, where $\bm{M}^v \in {\mathbb{R}^{N \times N}}$ reflects the manifold structure for the $v$th view. Taking LLE as an example, we could utilize $\bm{M}^v$ to reformulate the $\bm{{({I-W^v})}^T(I-W^v)}$. Thus, we could get the following equation for manifold learning problems:
 \begin{equation}
 \label{single view}
\begin{split}
&\mathop {\min }\limits_{{\bm{Y}^v}} tr(\bm{Y}^v \bm{M}^v \bm{Y}^{v^T}) \\
& s.t.  \quad {\bm{Y}^v}{\bm{C}^v}{\bm{Y}^v}^{^T} = \bm{I} \\
\end{split}
\end{equation}
 where $\bm{I}$ is an identity matrix and ${\bm{C}^v \in {\mathbb{R}^{N \times N}} }$ denotes the constraint term, which is a  square symmetric definite matrix. Taking LLE as an example, we could utilize $\bm{M}^v$ and $\bm{I}_N$to reformulate the ${({\bm{I}-\bm{W}^v})}^T(\bm{I}-\bm{W}^v)$ and  ${\bm{C}^v}$ respectively.

 For multi-view learning problems, multi-view data in $m$ views could be represented as $\bm{X}=\left\{ {\bm{X}^1,\bm{X}^{_2}, \ldots ,\bm{X}^m} \right\}$, where ${\bm{X}^v}=[ {\bm{x}_1^v,\bm{x}_2^v, \ldots ,\bm{x}_N^v} ] \in {\mathbb{R}^{D^v \times N}} $ denotes the features set with $N$ samples in the $v$th view and $D^v$ is the dimension of features set. By adding up objective function in Eq.(\ref{single view}) among all views, we could obtain the following optimization problem:
\begin{equation}\label{concat}
\begin{split}
&\mathop {\min }\limits_{{\bm{Y}^1},{\bm{Y}^2}, \ldots ,{\bm{Y}^m}} \sum\limits_{v = 1}^m tr(\bm{Y}^v \bm{M}^v \bm{Y}^{v^T})  \\
& s.t.  {\bm{Y}^v}{\bm{C}^v}{\bm{Y}^v}^{^T} = \bm{I},\forall 1 \le v \le m\ \\
\end{split}
\end{equation}
where ${\bm{Y}^v}=[ {\bm{y}_1^v,\bm{y}_2^v, \ldots ,\bm{y}_N^v} ] \in {\mathbb{R}^{d^v \times N}} $ denotes the low-dimensional embedding set with $N$ samples in the $v$th view and $\bm{d}^v$ is the dimension of features set such that $\bm{d}^v\ll\bm{D}^v$. But this equation is equal to solve the manifold learning based on single view problem for all views separately and fails to make full use of complementary and compatible information among multi-view features. Even though multiple views contain more information beyond single view, high dimension problems and integration among different views bring a huge challenge in obtaining the low-dimensional embedding. However, complementary and compatible information from multi-view features could greatly improve the performance of traditional methods. Therefore, it's wealthy and necessary to solve the above issues and the robust and scalable multi-view learning framework is proposed in this paper to obtain the low-dimensional embedding as  $\bm{Y}=\left\{ {\bm{Y}^1,\bm{Y}^{_2}, \ldots ,\bm{Y}^m} \right\}$.

\subsection{ Multi-view Framework based on Similarity Consensus}\label{main_framework}

On integrating different views, the dimension of the features set in each view owns its size, which is different from the other views. Besides, coordinating different manifold structures directly isn't easy to implement because of its intrinsic geometric properties in each view. Therefore, integrating different views is still full of challenges. To address two issues above, we make such a multi-view hypothesis that the pairwise similarity of coefficient vectors to be similar to all views. In this hypothesis, we encourage the pairwise similarities of instances to be similar across all views.

Considering the two-view case consisting of the $v$th view and the $w$th view. obviously, $\bm{Y}^v$ in the $v$th view and $\bm{Y}^w$ in the $w$th view have different dimensions $d^v$ and $d^w$, which are obtained respectively from different manifold structures. To handle these issues, we propose the following loss function as a measurement of consensus term between the $v$th view and the $w$th view:
\begin{equation}\label{reg0}
\begin{split}
&S\left( {{\bm{Y}^v},{\bm{Y}^w}} \right) = -\left\| {{\bm{K}^v} - {\bm{K}^w}} \right\|_F^2\\
\end{split}
\end{equation}
where $K^v$ and $K^v$ stand for the similarity matrix of the $v$th view and the $w$th view separately, and $\left\|  \cdot  \right\|_F^2$ denotes the square Frobienius norm (F-norm). To further express the consensus term, we expand Eq.(\ref{reg0}) as follows:
\begin{equation}\label{reg1}
\begin{split}
&S\left( {{\bm{Y}^v},{\bm{Y}^w}} \right) =  -\left\| {{\bm{K}^v} - {\bm{K}^w}} \right\|_F^2\\
& = -tr({(\bm{K}^v-\bm{K}^w)}^T(\bm{K}^v-\bm{K}^w))\\
& = -tr({\bm{K}^v}^T\bm{K}^v+{\bm{K}^w}^T\bm{K}^w-{\bm{K}^v}^T\bm{K}^w-{\bm{K}^w}^T\bm{K}^v)\\
\end{split}
\end{equation}
According the attributes of matrix trace and the symmetry of similarity matrix, we could reformulate Eq.(\ref{reg1}) as follows:
\begin{equation}
\begin{split}
&S\left( {{\bm{Y}^v},{\bm{Y}^w}} \right) = 2tr(\bm{K}^v\bm{K}^w)-tr(\bm{K}^v\bm{K}^v)-tr(\bm{K}^w\bm{K}^w)\\
\end{split}
\end{equation}
It's easy to find that the second term $tr(\bm{K}^v\bm{K}^v)$ and the third term $tr(\bm{K}^v\bm{K}^v)$ in the above equation just depend on individual view, which couldn't work in integrating two different views. Therefore, we could get the following equation by ignoring the irrelevant terms and the scaling term:
\begin{equation}\label{reg2}
\begin{split}
&S\left( {{\bm{Y}^v},{\bm{Y}^w}} \right) = tr(\bm{K}^v\bm{K}^w)\\
\end{split}
\end{equation}
To achieve the multi-view hypothesis, we want to maximize the above agreement between the $v$th view and the $w$th view. Combining the consensus term with Eq.(\ref{single view}), we could get the following joint optimization problem for the $v$th view and the $w$th view:
\begin{equation}
\begin{split}
&\mathop {\min }\limits_{{\bm{Y}^v},{\bm{Y}^w}} tr(\bm{Y}^v \bm{M}^v \bm{Y}^{v^T})+tr(\bm{Y}^w \bm{M}^w \bm{Y}^{w^T})-\lambda\bm{S}(\bm{Y}^v, \bm{Y}^w)\\
& s.t.\hspace{0.5em}{\bm{Y}^v}{\bm{C}^v}{\bm{Y}^v}^{^T} = \bm{I},{\bm{Y}^w}{\bm{C}^w}{\bm{Y}^w}^{^T} = \bm{I} \\
\end{split}
\end{equation}
where the hyperparameter $\lambda$ trades-off the above agreement function with the loss function in manifold learning under single view. We extend the two-view manifold learning framework for more than two views. It could be done by employing pairwise agreement in the objective function of Eq.(\ref{reg2}). Combing Eq.(\ref{reg2}) and Eq.(\ref{concat}), we have
\begin{equation}\label{pairwise_0}
\begin{split}
&\mathop {\min }\limits_{{\bm{Y}^1},{\bm{Y}^2}, \ldots ,{\bm{Y}^m}} \sum\limits_{v = 1}^m tr(\bm{Y}^v \bm{M}^v \bm{Y}^{v^T}) -\lambda \sum\limits_{1 \le v \ne w \le m} {S({Y^v},{Y^w})} \\
& s.t. {\bm{Y}^v}{\bm{C}^v}{\bm{Y}^v}^{^T} = \bm{I},\forall 1 \le v \le m\ \\
\end{split}
\end{equation}
Inspired with these works \cite{xia2010multiview, xu2017re, nie2017auto}, we realize that different view plays different importance in learning to obtain the low-dimensional embedding. Thus, we allocate the different weights for different views, which reflect the importance that each view plays in learning to obtain the low-dimensional embedding. Combining this with Eq.(\ref{pairwise_0}) for $m$ views, we obtain the following overall objective function of multi-view manifold learning framework:
\begin{equation}\label{pairwise}
\begin{split}
&\mathop {\min }\mathcal{F}({\bm{Y}^1},{\bm{Y}^2}, \ldots ,{\bm{Y}^m},\bm{\alpha}) = \\
&\sum\limits_{v = 1}^m {(\bm{\alpha}^v)}^r tr(\bm{Y}^v \bm{M}^v \bm{Y}^{v^T})-\lambda \sum\limits_{1 \le v \ne w \le m} {S({Y^v},{Y^w})} + \gamma{\Vert\bm{\alpha}\Vert}_r^r\\
& s.t. {\bm{Y}^v}{\bm{C}^v}{\bm{Y}^v}^{^T} = \bm{I},{\bm{\alpha} ^v}>0, \forall 1 \le v \le m \\
& \quad \sum\limits_{v = 1}^m {{\bm{\alpha} ^v}}  = 1 \\
\end{split}
\end{equation}
where $\bm{\alpha}  = \left[ {{\bm{\alpha} ^1},{\bm{\alpha} ^2}, \cdots ,{\bm{\alpha} ^m}} \right]$ is a non-negative weights vector, $r>1$ is scalar controlling the weights and $\gamma$ is coefficient of regular term for $\bm{\alpha}$. The first term ensures that the manifold structure would be maintained in low dimensional space, the second term follows the multi-view hypothesis by pairwise consensus and the third constrains the scale of weights to improve robustness of our methods. In this way, we obtain a multi-view manifold learning framework based on pairwise consensus.

Although multi-view features extract from different means, they are just different descriptions for one same object. Therefore, considering that all manifold spaces of different are integrated into one common manifold space is also a Feasible idea. And we utilize the common space as the centroid view. Accordingly, we propose such a multi-view hypothesis that the similarties between the $v$th view and centroid view are similar across all views. This hypothesis means that all similarity matrices from multiple views should be consistent with the similarity of the centroid view. To achieve this multi-view hypothesis, we want to maximize the above agreement between the $v$th view and the centroid view. Combining this with Eq.(\ref{pairwise}), the objective function of multi-view manifold learning framework based on the centroid view could be written as follows:
\begin{equation}\label{centroid}
\begin{split}
&\mathop {\min }\mathcal{F}({\bm{Y}^*},{\bm{Y}^1},{\bm{Y}^2}, \ldots ,{\bm{Y}^m},\bm{\alpha}) = \\
&\sum\limits_{v = 1}^m {(\bm{\alpha}^v)}^r tr(\bm{Y}^v \bm{M}^v \bm{Y}^{v^T}) - \lambda \sum\limits_{1 \le v \le m} {\bm{S}({\bm{Y}^v},{\bm{Y}^*})} + \gamma{\Vert\bm{\alpha}\Vert}_r^r\\
& s.t. {\bm{Y}^*}{\bm{Y}^*}^{^T} = \bm{I}, {\bm{Y}^v}{\bm{C}^v}{\bm{Y}^v}^{^T} = \bm{I}, {\bm{\alpha} ^v}>0, \forall 1 \le v \le m \\
& \quad \sum\limits_{v = 1}^m {{\bm{\alpha} ^v}}  = 1\\
\end{split}
\end{equation}
where ${\bm{Y}^*}=[ {\bm{y}_1^*,\bm{y}_2^*, \ldots ,\bm{y}_N^*} ] \in {\mathbb{R}^{d^* \times N}} $ denotes the low-dimensional embedding and $\bm{d}^*$ is the dimension of the centroid view. In contrast to the pairwise agreement strategy which has $m(m-1)/2$ pairwise agreement terms, the centroid-based strategy has $m$ pairwise agreement terms.

In summary, we derive two multi-view manifold learning frameworks based on pairwise consensus and centroid consensus separately in this section, which encourage all similarity matrices from multiple views to be consistent with each other. The first scheme enforces that each view pair $(v,w)$ should have high pairwise similarity by the consensus term $S({Y^v},{Y^w})$ and the second scheme enforces each view to maintain similar by integrating into one common manifold space. To solve Eq.(\ref{pairwise}) and Eq.(\ref{centroid}), we propose an alternating optimization algorithm with guaranteed convergence, as shown in the next section.

\subsection{Optimization}\label{optimization}
In this section,  we derive the solutions for the above problems, which are nonlinearly constrained nonconvex optimization problems. To the best of our knowledge, there is no direct way to get the global optimal solution. For this reason, we propose an iterative alternating strategy based on the alternating optimization\cite{bezdek2002some} to obtain a local optimal solution. That is to say, we alternatively update each variable when fixing others.

$\bm{Updating \quad \bm{\alpha}^v}$: When we fix other variables but $\bm{\alpha}^v$, both the Eq.(\ref{pairwise}) and the Eq.(\ref{centroid}) w.r.t $\bm{\alpha}^v$ could be written as follows:
\begin{equation}\label{solve_alpha}
\begin{split}
&\mathop {\min } \mathcal{F}({{\bm{\alpha}^1},{\bm{\alpha}^2}, \ldots ,{\bm{\alpha}^m}}) =\sum\limits_{v = 1}^m {(\bm{\alpha}^v)}^r tr(\bm{Y}^v \bm{M}^v \bm{Y}^{v^T}) + \gamma{\Vert\bm{\alpha}\Vert}_r^r\\
& s.t. \sum\limits_{v = 1}^m {{\bm{\alpha} ^v}}  = 1, \bm{\alpha} ^v>0, \forall 1 \le v \le m \\
\end{split}
\end{equation}
which could be solved by the method of Lagrange multiplier. BY introducing the Lagrange multiplier $\eta$, the above problem could be transformed as follows:
\begin{equation}
\begin{split}
&\mathop {\min } \mathcal{L}({{\bm{\alpha}^1},{\bm{\alpha}^2}, \ldots ,{\bm{\alpha}^m}, \eta}) =\sum\limits_{v = 1}^m {(\bm{\alpha}^v)}^r tr(\bm{Y}^v \bm{M}^v \bm{Y}^{v^T}) \\
& \quad \quad + \gamma{\Vert\bm{\alpha}\Vert}_r^r -\eta(\sum\limits_{v = 1}^m {{\bm{\alpha} ^v}}-1) \\
\end{split}
\end{equation}
Its partial derivatives with respect to $\bm{\alpha}^v$ and $\eta$ are:
\begin{equation}
\left\{
\begin{array}{l}
\frac{\partial\mathcal{L}}{\partial \bm{\alpha} ^v} = r{(\bm{\alpha}^v)}^{r-1}tr(\bm{Y}^v \bm{M}^v \bm{Y}^{v^T})+ r{(\bm{\alpha}^v)}^{r-1}\gamma-\eta \\
\frac{\partial\mathcal{L}}{\partial \eta} = \sum\limits_{v = 1}^m {{\bm{\alpha} ^v}}-1
\end{array}
\right.
\end{equation}
By setting $\bigtriangledown_{\bm{\alpha}^v,\eta}\mathcal{L}=0$, we get the optimal solution of $\bm{\alpha}^v$ as:
\begin{equation}
\begin{split}
\bm{\alpha}^v=\frac{{(1/(tr(\bm{Y}^v \bm{M}^v \bm{Y}^{v^T})+\gamma))}^{1/(r-1)}}{\sum\limits_{w = 1}^m {{(1/(tr(\bm{Y}^w \bm{M}^w \bm{Y}^{w^T})+\gamma))}^{1/(r-1)}}}
\end{split}
\end{equation}

$\bm{Updating \quad \bm{Y}^v}$: Because of the difference between dependent terms solving $\bm{Y}^v$, we solve the optimal solution $\bm{Y}^v$ in the Eq.(\ref{pairwise}) and the Eq.(\ref{centroid}) respectively. For the Eq.(\ref{pairwise}), by removing the irrelevant terms, we obtain the following problem:
\begin{equation}\label{pairwise_1}
\begin{split}
&\mathop {\min }\mathcal{F}({{\bm{Y}^v}}) = {(\bm{\alpha}^v)}^r tr(\bm{Y}^v \bm{M}^v \bm{Y}^{v^T}) -\lambda \sum\limits_{1 \le w \ne v \le m} {S({Y^v},{Y^w})} \\
& s.t.  \quad {\bm{Y}^v}{\bm{C}^v}{\bm{Y}^v}^{^T} = \bm{I}\ \\
\end{split}
\end{equation}
To get a conveniently solved optimization problem influenced by the form of Eq.(\ref{single view}), we choose linear kernel as kernel function, i.e., $k^v(x_i^v, x_j^v)={(x_i^v)}^Tx_j^v$. Besides, we could take care of the nonlinearity present in the $v$th view by choosing the suitable kernel function in other views. In this way, we reformulate the Eq.(\ref{pairwise_1}) to
\begin{equation}\label{pairwise_2}
\begin{split}
&\mathop {\min }\mathcal{F}({{\bm{Y}^v}}) = {(\bm{\alpha}^v)}^r tr(\bm{Y}^v \bm{M}^v \bm{Y}^{v^T}) -\lambda \sum\limits_{1 \le w \ne v \le m} {tr(\bm{Y}^v \bm{K}^w \bm{Y}^{v^T})} \\
& s.t.  \quad {\bm{Y}^v}{\bm{C}^v}{\bm{Y}^v}^{^T} = \bm{I}\ \\
\end{split}
\end{equation}
 Based on the Ky-Fan theory\cite{bhatia2013matrix}, $\bm{Y}^v$ in Eq.(\ref{pairwise_2}) has a global optimal solution, which is given as the eigenvectors associated with the smallest $d^v$ eigenvalues of ${(\bm{\alpha}^v)}^r\bm{M}^v -\lambda \sum\limits_{1 \le w \ne v \le m} \bm{K}^w$.
We set $\bm{L}^v={(\bm{\alpha}^v)}^r\bm{M}^v -\lambda \sum\limits_{1 \le w \ne v \le m} \bm{K}^w$. It's easy to find that $\bm{L}^v$ is symmetric. Applying Lagrangian multiplier method to solve Eq.(\ref{centroid_2}), we could get $\bm{L}^v\bm{Y}^v=\eta\bm{C}^v\bm{Y}^v$, where $\eta$ is the multiplier coefficient. Because $\bm{C}^v$ is  a  square symmetric definite matrix, the optimal solution of $\bm{Y}^v$ in Eq.(\ref{centroid_2}) is the eigenvectors associated with the smallest $d^v$ eigenvalues of ${(\bm{C}^v)}^{-1}\bm{L}^v$.

For the Eq.(\ref{centroid}), by removing the irrelevant terms, we obtain the following problem:
\begin{equation}\label{centroid_1}
\begin{split}
&\mathop {\min }\mathcal{F}({\bm{Y}^v}) =\bm{\alpha}^v tr(\bm{Y}^v \bm{M}^v \bm{Y}^{v^T}) -\lambda  {S({Y^v},{Y^*})} \\
& s.t.  \quad {\bm{Y}^v}{\bm{C}^v}{\bm{Y}^v}^{^T} = \bm{I}\ \\
\end{split}
\end{equation}
We also formulate kernel function as $k^v(x_i^v, x_j^v)={(x_i^v)}^Tx_j^v$ according to the previous derivation process. And we could reformulate the Eq.(\ref{centroid_1}) as
\begin{equation}\label{centroid_2}
\begin{split}
&\mathop {\min }\mathcal{F}({\bm{Y}^v})= {(\bm{\alpha}^v)}^r tr(\bm{Y}^v \bm{M}^v \bm{Y}^{v^T}) -\lambda {tr(\bm{Y}^v \bm{K}^* \bm{Y}^{v^T})} \\
& s.t.  \quad {\bm{Y}^v}{\bm{C}^v}{\bm{Y}^v}^{^T} = \bm{I}\ \\
\end{split}
\end{equation}
We set $\bm{L}^v={(\bm{\alpha}^v)}^r\bm{M}^v -\lambda \bm{K}^*$. Obviously, $\bm{L}^v$ is also symmetric. Applying Lagrangian multiplier method to solve Eq.(\ref{centroid_2}), we could get $\bm{L}^v\bm{Y}^v=\eta\bm{C}^v\bm{Y}^v$, where $\eta$ is the multiplier coefficient. Because $\bm{C}^v$ is  a  square symmetric definite matrix, the optimal solution of $\bm{Y}^v$ in Eq.(\ref{centroid_2}) is the eigenvectors associated with the smallest $d^v$ eigenvalues of ${(\bm{C}^v)}^{-1}\bm{L}^v$.

$\bm{Updating} \quad \bm{Y}^*$:  When we fix other variables but $\bm{Y}*$ in the Eq.(\ref{centroid}), the Eq.(\ref{centroid}) w.r.t $\bm{Y}*$ could be written as follows:
\begin{equation}\label{centroid_*}
\begin{split}
&\mathop {\min }\mathcal{F}({\bm{Y}^*}) = -\lambda \sum\limits_{1 \le v \le m} {S({Y^v},{Y^*})} \\
& s.t.  \quad {\bm{Y}^*}{\bm{Y}^*}^{^T} = \bm{I}\ \\
\end{split}
\end{equation}
Here, inspired by the previous section, we choose the linear kernel as kernel function for the common view. Substituting this into the Eq.(\ref{centroid_*}) and neglecting the scale term, then we could obtain the following optimization problem:
\begin{equation}\label{centroid_final}
\begin{split}
&\mathop {\max }\mathcal{F}({\bm{Y}^*}) = \sum\limits_{1 \le v \le m} {tr(\bm{Y}^* \bm{K}^v \bm{Y}^{*^T})} \\
& s.t.  \quad {\bm{Y}^*}{\bm{Y}^*}^{^T} = \bm{I} \\
\end{split}
\end{equation}
We find that $ \sum\limits_{1 \le v \le m} \bm{K}^v$ is symmetric. According to the Ky-Fan theory\cite{bhatia2013matrix}, the optimal solution of $\bm{Y}^*$ in Eq.(\ref{centroid_final}) is the eigenvectors associated with the largest $d^*$ eigenvalues of $ \sum\limits_{1 \le v \le m} \bm{K}^v$.

We have so far presented the whole optimization procedures for the Eq.(\ref{pairwise}) and the Eq.(\ref{centroid}). According to the descriptions above, we can form the alternating optimization strategy to iteratively update the variables in Eq.(\ref{pairwise}) and Eq.(\ref{centroid}), summarized in \textbf{Algorithm 1} and \textbf{Algorithm 2}, to capture a local optimal solution of our framework.

\begin{algorithm}
\caption{The optimization procedure of Pairwise Consensus-based Multi-view Learning Framework}
\label{Pairwise Framework}
\hspace*{0.02in} {\bf Input:}

\hspace*{0.05in}1. A multi-view features set with N training samples having m views ${X^v} = [x_1^v,x_2^v, \ldots ,x_N^v] \in {\mathbb{R}^{{D_v} \times N}}, v=1,2,\dots,m$.

\hspace*{0.05in}2. The hyperparameter parameter $\gamma$ in Eq.(\ref{pairwise}).

\hspace*{0.02in} {\bf Output:}\hspace*{0.08in}The embedding $[Y^1,Y^2, \ldots, Y^m]$  of m views.

\hspace*{0.02in} {\bf The Main Procedure:}
\begin{algorithmic}
\FOR{v=1:m}
    \STATE{3. Initialize $\bm{\alpha}^v=1/m$.}
    \STATE{4. Specialize the $M^v$ in Eq.(\ref{pairwise_2})}
    \STATE{5. Initialize $\bm{Y}^v$ according to $M^v$}
\ENDFOR
\REPEAT
\FOR{v=1:m}
    \STATE{6. Update $Y^v$ for the $v$th view by solving Eq.(\ref{pairwise_2}).}
\ENDFOR
\STATE{7. Update $\bm{\alpha}$ by solving Eq.(\ref{solve_alpha}).}
\UNTIL{$[Y^1,Y^2, \ldots, Y^m]$ converges}
\end{algorithmic}
\end{algorithm}

\begin{algorithm}
\caption{The optimization procedure of Centroid Consensus-based Multi-view Learning Framework}
\label{Centroid Framework}
\hspace*{0.02in} {\bf Input:}

\hspace*{0.05in}1. A multi-view features set with N training samples having m views ${X^v} = [x_1^v,x_2^v, \ldots ,x_N^v] \in {\mathbb{R}^{{D_v} \times N}}, v=1,2,\dots,m$.

\hspace*{0.05in}2. The hyperparameter parameter $\gamma$ in Eq.(\ref{centroid}).

\hspace*{0.02in} {\bf Output:}\hspace*{0.08in}The centroid embedding $Y^*$

\hspace*{0.02in} {\bf The Main Procedure:}
\begin{algorithmic}
\FOR{v=1:m}
    \STATE{3. Initialize $\bm{\alpha}^v=1/m$.}
    \STATE{4. Specialize the $M^v$ in Eq.(\ref{centroid_2})}
    \STATE{5. Initialize $\bm{Y}^v$ according to $M^v$}
\ENDFOR
\REPEAT
\STATE{6. Update $Y^*$ by solving Eq.(\ref{centroid_final})}
\FOR{v=1:m}
    \STATE{7. Update $Y^v$ for the $v$th view by solving Eq.(\ref{centroid_2}).}
\ENDFOR
\STATE{8. Update $\bm{\alpha}$ by solving Eq.(\ref{solve_alpha}).}
\UNTIL{$Y^*$ converges}
\end{algorithmic}
\end{algorithm}

Because our proposed framework is solved by alternating optimization strategy, it's essential to discuss the convergence of \textbf{Algorithm 1} and \textbf{Algorithm 2}. For \textbf{Algorithm 1}, it's obvious that \{${\bm{Y}^1},{\bm{Y}^2}, \ldots ,{\bm{Y}^m},\bm{\alpha}$\} generated via solving Eq.(\ref{pairwise_2}) and Eq.(\ref{solve_alpha}) are the exact minimum points of Eq.(\ref{pairwise_2}) and Eq.(\ref{solve_alpha}) respectively. As a result, the value of the objective function $\mathbb{F}({\bm{Y}^1},{\bm{Y}^2}, \ldots ,{\bm{Y}^m},\bm{\alpha})$ in Eq.(\ref{pairwise}) is decreasing in each iteration of \textbf{Algorithm 1}. Thus, the alternating optimization will monotonically decrease the objective in Eq.(\ref{pairwise}) until it convergences. As the above discussion, the objective in Eq.(\ref{pairwise}) could convergence a stationary point. Similarly, we also show the convergence of \textbf{Algorithm 2} according to the above analysis process of \textbf{Algorithm 1}.

\subsection{Time Complexity Analysis}
To clearly explain the efficiency of our framework, this subsection mainly analyzes the influences on time complexity. The computational cost mainly consists of two parts. The first part is initialization for all variables, which mainly depends on the construction for $\bm{M}^v$ and performed singular value decomposition for $\bm{Y}^v$. The second part is for low-dimensional embedding $\bm{Y}^v$ or $\bm{Y}^*$, which need to perform eigenspace decomposition in each iteration. Therefore, the time complexity of \textbf{Algorithm 1} and \textbf{Algorithm 2} process scales are  O($mN^3+T(m+1)N^3$) and O($mN^3+T(m+1)N^3$) respectively, approximately O($TmN^3$), where $T$ is the iteration times of the alternating optimization procedure, $m$ is the number of views and $N$ is the number of samples. When comparing with DR methods under single view, the time consumption of our framework is higher due to the continuous iterative optimization process. But when comparing with those multi-view methods which employ eigenspace decomposition and iterative alternating strategy, such as MSE and Co-regularized whose time complexity scales is about O($TmN^3$), our framework is quite comparable to these methods.

\section{Extensions}
In this section, we extend subspace learning and kernel learning methods based on similarity consensus into the multi-view framework. Most subspace learning methods could be cast as a special form of QCQP. Specially, the optimal projection $\bm{W}^v$ for the $v$th view could be obtained as
\begin{equation}\label{subspace_learning}
\begin{split}
&\mathop{\max }\limits_{\bm{w}^v} \hspace{0.5em} tr(\bm{w}^{v^T}\bm{A}^v\bm{w}^v)\\
& s.t. \quad \bm{w}^{v^T}\bm{B}^v\bm{w^v} = \bm{I}\\
\end{split}
\end{equation}
where $\bm{W}^v \in \mathbb{R}^{D^v \times d^v}$ denotes the projection matrix for the $v$th view, $\bm{A}^v$ is some symmetric square matrix and $\bm{B}^v$ is a square symmetric definite matrix. Methods that fit this equation include PCA\cite{wold1987principal}, LDA\cite{mika1999fisher}, LPP\cite{he2004locality} and MFA\cite{xu2007marginal}.

\subsection{Multi-view Subspace Learning based on Similarity Consensus}
Most existing subspace learning methods are proposed to find a low dimensional space by preserving come properties of raw features to significantly reduce time consuming and computational cost for high dimensional features. However, these DR methods above mainly focus on single view so that information among multiple views couldn't be fully utilized. Thus, we extend those methods based on single view to our multi-view framework. For the $v$th view, we could denote $\bm{W}^{v^T}\bm{X}^v$ as $\bm{Y}^v$ when we use $\bm{W}^v$ as the projection matrix. Referring to the construction process of Eq.(\ref{pairwise}), we further factorize $\bm{A}^v$ and $\bm{B}^v$ as $\bm{X}^v\bm{M}\bm{X}^{v^T}$ and $\bm{X}^v\bm{C}^v\bm{X}^{v^T}$ respectively. In this way, it's more convenient to explain the expansion process. Taking PCA, NPE and LDA as examples, we could express $\bm{M}^v$ and $\bm{C}^v$ as follows:
\begin{itemize}
    \item PCA $\bm{M}^v={(\bm{I}_N-\bm{1}_{1/N})}^T(\bm{I}_N-\bm{1}_{1/N})$, where $\bm{I}_N \in \mathbb{R}^{N \times N}$ is the identity matrix and $\bm{1}_{1/N}$ is a  $\mathbb{R}^{d_v \times d_v}$ matrix that each element is filled with $1/N$. And $\bm{C}^v={(\bm{X}^{v^T}\bm{X}^v)}^{-1}$ is a $\mathbb{R}^{d_v \times d_v}$ matrix.
    \item NPE $\bm{M}^v=-{({\bm{I}_N-\bm{W}^v})}^T(\bm{I}_N-\bm{W}^v)$, where $\bm{W}^v \in \mathbb{R}^{N \times N}$ is the reconstruction coefficient matrix in the $v$th view. And $\bm{C}^v=\bm{I}_N$, where $\bm{I}_N \in \mathbb{R}^{N \times N}$ is the identity matrix.
    \item LDA $\bm{M}_{i,j}^v=1/N_v^c$ if $\bm{X}_i^v$ and $\bm{X}_j^v$ belong to the class $c$, 0 otherwise, where $N_v^c$ is the number of samples for class $c$ in the $v$th view. And $\bm{C}^v=\bm{I}_N-\bm{M}^v$,  where $\bm{I}_N \in \mathbb{R}^{N \times N}$ is the identity matrix.
\end{itemize}
Combing this with Eq.(\ref{pairwise}), we could get the following pairwise based multi-view subspace learning optimization problem:
\begin{equation}\label{pairwise_subspace}
\begin{split}
&\mathop {\max }\mathcal{F}(\bm{\alpha}, {\bm{W}^1},{\bm{W}^2}, \ldots ,{\bm{W}^m}) =\\
&\sum\limits_{v = 1}^m {(\bm{\alpha}^v)}^r tr(\bm{W}^{v^T}\bm{X}^v\bm{M}^v\bm{X}^{v^T}\bm{W^v})-\gamma{\Vert\bm{\alpha}\Vert}_r^r + \\
&\gamma \sum\limits_{1 \le v \ne w \le m} {S(\bm{W}^{v^T}\bm{X}^v,\bm{W}^{w^T}\bm{X}^w)} \\
& s.t. \bm{W}^{v^T}\bm{X}^v\bm{B}^v\bm{X}^{v^T}\bm{W^v} = \bm{I},\forall 1 \le v \le m \\
& \quad \sum\limits_{v = 1}^m {{\bm{\alpha} ^v}}  = 1\\
\end{split}
\end{equation}
Similarly, we also obtain the following centroid consensus-based multi-view subspace learning optimization problem:
\begin{equation}\label{centroid_subspace}
\begin{split}
&\mathop {\max }\mathcal{F}(\bm{\alpha}, {\bm{W}^1},{\bm{W}^2}, \ldots ,{\bm{W}^m}, {\bm{Y}^*}) = \\
&\sum\limits_{v = 1}^m {(\bm{\alpha}^v)}^r tr(\bm{W}^{v^T}\bm{X}^v\bm{M}^v\bm{X}^{v^T}\bm{W^v}) - \gamma{\Vert\bm{\alpha}\Vert}_r^r+ \\
&\gamma \sum\limits_{1 \le v \le m} {S(\bm{W}^{v^T}\bm{X}^v,\bm{Y}^*)} \\
& s.t. {\bm{Y}^*}{\bm{Y}^*}^{^T} = \bm{I}, \bm{W}^{v^T}\bm{X}^v\bm{B}^v\bm{X}^{v^T}\bm{W^v} = \bm{I},\forall 1 \le v \le m \\
& \quad \sum\limits_{v = 1}^m {{\bm{\alpha} ^v}}  = 1\\
\end{split}
\end{equation}
Reforing to \textbf{Algorithm 1} and \textbf{Algorithm 2}, we could obtain the optimal solution of Eq.(\ref{pairwise_subspace}) and Eq.(\ref{centroid_subspace}). Through the above construction process, we extend those methods based on single view to our multi-view framework.

\subsection{Multi-view Kernel Learning based on Similarity Consensus}
Kernel methods\cite{scholkopf1997kernel, mika1999fisher, torresani2007large} involve mapping to non-linear space and then obtain projection in that mapped space. Specially, for the $\bm{X}^v$ in the $v$th view, we first map all features into the kernel space by nonlinear function $\phi^v()$, i.e.$\bm{X}_{\phi}^v=[\phi^v(\bm{x}_1^v), \phi^v(\bm{x}_2^v), \ldots, \phi^v(\bm{x}_N^v)]$. And it has been verified \cite{scholkopf1997kernel} that $\bm{W}_{\phi}^v$ is that mapped space spanned by $\phi^v(\bm{x}_1^v), \phi^v(\bm{x}_2^v), \ldots, \phi^v(\bm{x}_N^v)$. Therefore, $\bm{W}_{\phi}^v$ could be expressed as follows:
\begin{equation}
    \begin{split}
        \bm{W}_{\phi}^v=\sum\limits_{i = 1}^N {\phi^v(\bm{x}_i^v)\bm{\beta}_i^v}=\bm{X}_{\phi}^v\bm{\beta}^v
    \end{split}
\end{equation}
 where $\bm{\beta}^v={[\bm{\beta}_1^v, \bm{\beta}_2^v, \ldots, \bm{\beta}_N^v]}^T \in \mathbb{R}^{N \times d^v}$ consists of the expansion coefficients. Combing this with Eq.(\ref{pairwise_subspace}), introduced in the previous section we could get the following pairwise based multi-view kernel learning optimization problem:
 \begin{equation}\label{pairwise_kernel_0}
\begin{split}
&\mathop {\max }\mathcal{F}(\bm{\alpha}, {\bm{\beta}^1},{\bm{\beta}^2}, \ldots ,{\bm{\beta}^m}) = \\
&\sum\limits_{v = 1}^m {(\bm{\alpha}^v)}^r tr(\bm{\beta}^{v^T}{\phi^v(\bm{x}_N^v)}^T \phi^v(\bm{x}_N^v)\bm{M}^v{\phi^v(\bm{x}_N^v)}^T \phi^v(\bm{x}_N^v) \bm{\beta}^v)- \gamma{\Vert\bm{\alpha}\Vert}_r^r+\\
&\gamma \sum\limits_{1 \le v \ne w \le m} {S(\bm{\beta}^{v^T}\bm{K}^v,\bm{\beta}^{w^T}\bm{K}^w)}  \\
& s.t. \bm{\beta}^{v^T}{\phi^v(\bm{x}_N^v)}^T \phi^v(\bm{x}_N^v)\bm{C}^v {\phi^v(\bm{x}_N^v)}^T \phi^v(\bm{x}_N^v) \bm{\beta}^v= \bm{I},\forall 1 \le v \le m \\
& \quad \sum\limits_{v = 1}^m {{\bm{\alpha} ^v}}  = 1\\
\end{split}
\end{equation}

We set $\bm{K}_{\phi}^v={\phi^v(\bm{x}_N^v)}^T \phi^v(\bm{x}_N^v)$. Substituting this in Eq.(\ref{pairwise_kernel_0}), we could further rewrite Eq.(\ref{pairwise_kernel_0}) as follows:
\begin{equation}\label{pairwise_kernel_1}
\begin{split}
&\mathop {\max }\mathcal{F}(\bm{\alpha}, {\bm{\beta}^1},{\bm{\beta}^2}, \ldots ,{\bm{\beta}^m}) = \\
&\sum\limits_{v = 1}^m {(\bm{\alpha}^v)}^r tr(\bm{\beta}^{v^T}\bm{K}^v\bm{M}^v\bm{K}^v \bm{\beta}^v)- \gamma{\Vert\bm{\alpha}\Vert}_r^r+\\
&\gamma \sum\limits_{1 \le v \ne w \le m} {S(\bm{\beta}^{v^T}\bm{K}^v,\bm{\beta}^{w^T}\bm{K}^w)}  \\
& s.t. \bm{\beta}^{v^T} \bm{K}_{\phi}^v \bm{C}^v \bm{K}_{\phi}^v\bm{\beta}^v = \bm{I},\forall 1 \le v \le m \\
& \sum\limits_{v = 1}^m {{\bm{\alpha} ^v}}  = 1\\
\end{split}
\end{equation}
Similarly, we also obtain the following centroid based multi-view subspace learning optimization problem:
\begin{equation}\label{centroid_kernel}
\begin{split}
&\mathop {\max }\mathcal{F}(\bm{\alpha}, {\bm{\beta}^1},{\bm{\beta}^2}, \ldots ,{\bm{\beta}^m}, {\bm{Y}^*}) =  \\
&\sum\limits_{v = 1}^m {(\bm{\alpha}^v)}^r tr(\bm{\beta}^{v^T}\bm{K}^v\bm{M}^v \bm{K}^v \bm{\beta}^v)-\gamma{\Vert\bm{\alpha}\Vert}_r^r+ \\
&\gamma \sum\limits_{1 \le v \le m} {S(\bm{\beta}^{v^T}\bm{K}^v,\bm{Y}^*)} \\
& s.t.  {\bm{Y}^*}{\bm{Y}^*}^{^T} = \bm{I}, \bm{\beta}^{v^T} \bm{K}^v \bm{C}^v \bm{K}^v\bm{\beta}^v = \bm{I},\forall 1 \le v \le m \\
& \quad \sum\limits_{v = 1}^m {{\bm{\alpha} ^v}}  = 1\\
\end{split}
\end{equation}
Reforing to \textbf{Algorithm 1} and \textbf{Algorithm 2}, we could obtain the optimal solution of Eq.(\ref{pairwise_kernel_1}) and Eq.(\ref{centroid_kernel}) by the iterative alternating strategy. Through the above construction process, we extend kernel methods based on single view to our multi-view framework.

\section{Experiments}
In this section, we evaluate the performance of our framework by comparing it with several classical DR methods and multi-view learning methods in the multi-view datasets of texts and images. These experiment results verify the excellent performance of our framework.

\begin{figure*}[htbp]
\centering
\begin{minipage}{0.55\textwidth}
\centerline{\includegraphics[width=0.95\textwidth]{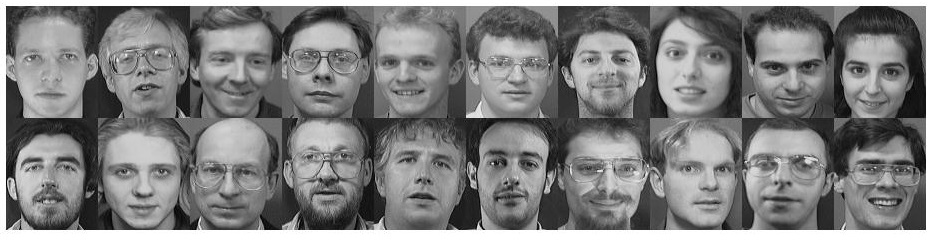}}
\centerline{(a) Some images in ORL dataset}
\vfill
\centerline{\includegraphics[width=0.95\textwidth]{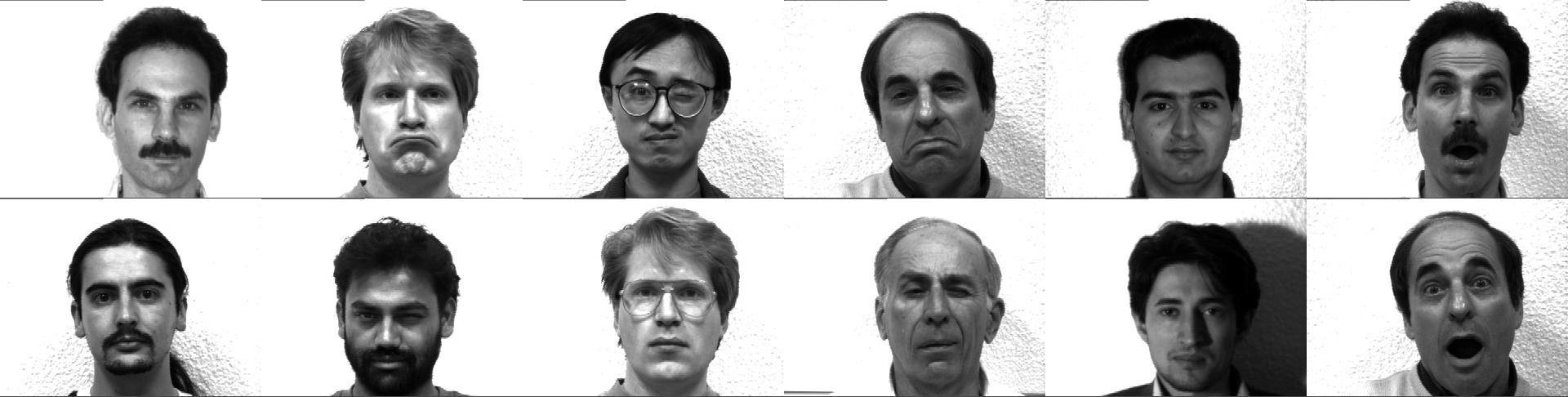}}
\centerline{(b) Some images in Yale dataset}
\end{minipage}
\hfill
\begin{minipage}{0.42\linewidth}
\centerline{\includegraphics[width=0.9\textwidth]{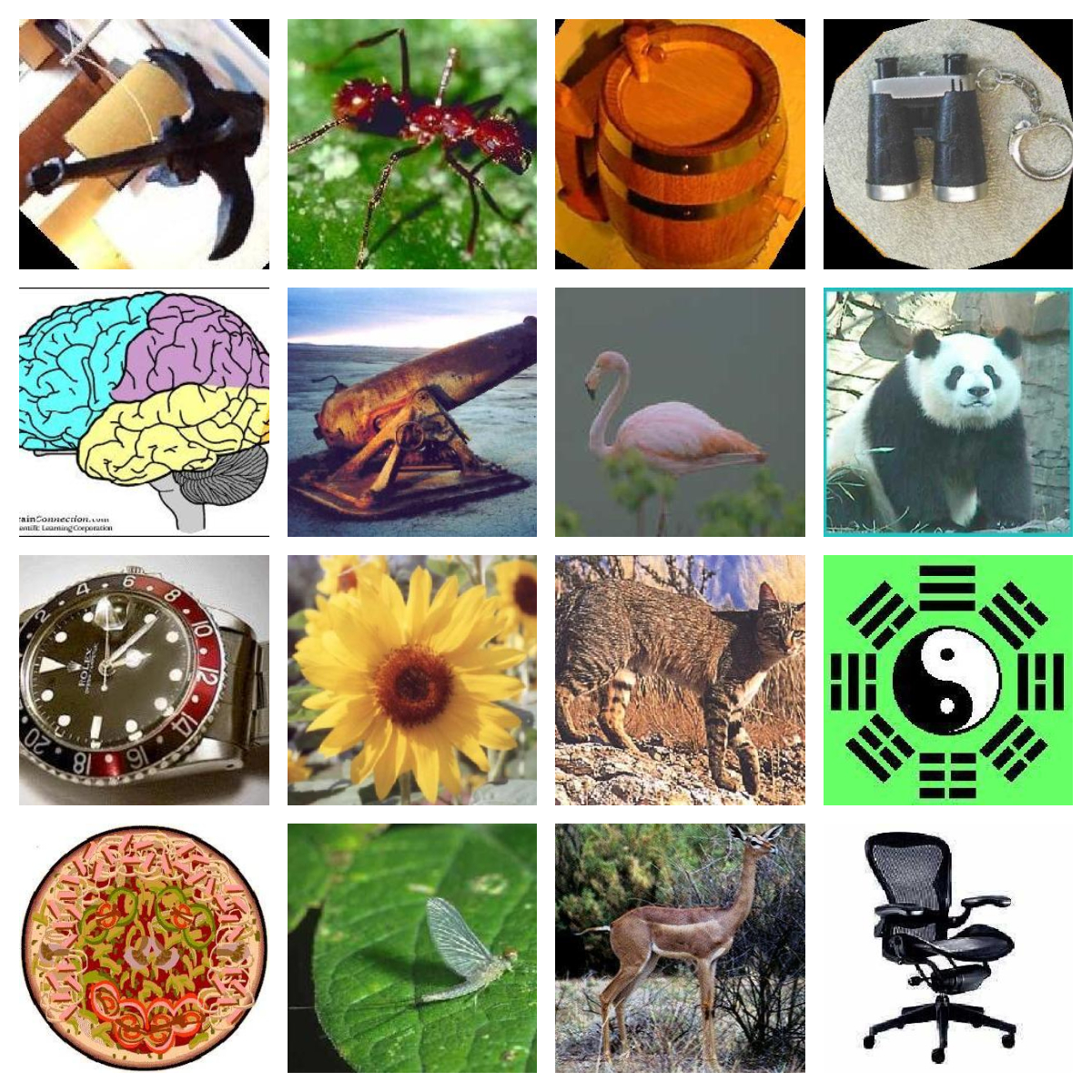}}
\centerline{(c) Some images in Caltech101 dataset}
\end{minipage}

\caption{Examples Images}
\label{example_images}
\end{figure*}

\subsection{Datasets and Competitors}
There are six datasets in the form of texts and images. Three text datasets adopted in the experiments are widely used in works, including 3Source\footnotetext[1]{http://mlg.ucd.ie/datasets/3sources.html} and Cora\footnotetext[2]{3http://lig-membres.imag.fr/grimal/data.html}. 3Sources consist of 3 well-known online news sources: BBC, Reuters and the Guardian, and each source is treated as one view. We select the 169 stories which are reported in all these 3 sources; Cora consists of 2708 scientific publications which come from 7 classes. Because the document is represented by content and cites views, Cora could be considered as a two views datasets. Three images datasets adopted in the experiments are widely used in works, including: ORL\footnotetext[3]{http://www.uk.research.att.com/facedatabase.html}, Yale\footnotetext[4]{http://cvc.yale.edu/projects/yalefaces/yalefaces.html}, Caltech 101\footnotetext[5]{http://www.vision.caltech.edu/ImageDatasets/Caltech101/}.
ORL and Yale are two face image datasets which have been widely used in face recognition. Caltech101 is a benchmark image dataset which contains 9144 images corresponding to 102 objects. We extract features for images using three different image descriptors. The detailed information of these datasets is summarized in Table \ref{datasets}. Some example images in YALE and ORL datasets are shown in the Fig.\ref{example_images}.

\begin{table}[htbp]
\caption{The detail information of the multi-view datasets}
\label{datasets}
\centering
\begin{tabular*}{0.45\textwidth}{@{\extracolsep{\fill}}llll}  
\hline
Datasets &Samples &Classes &Views\\
\hline  
Sources &169 &6 &3\\
Cora &2708 &7 &2\\
ORL &400 &40 &3\\
Yale &165 &15 &3\\
Caltech101 &9144 &102 &3\\
\hline
\end{tabular*}
\end{table}

We propose a multi-view framework for manifold learning in this paper, and extend subspace learning and kernel learning methods to this framework. Here, we choose two multi-view manifold learning methods implemented in our framework, which are centroid-based multi-view local linear embedding(CMLLE) and pairwise-based multi-view laplacian embedding(PMLE) separately. The effectiveness of our framework is evaluated by comparing CMLLE and PMLE with the following algorithms, including: the best performance of the single view-based LLE(BLLE), the best performance of the single view-based LE(BLE), the feature concatenation-based LLE(CLLE), MSE,CCA. All methods are evaluated 30 times with different random training samples and testing samples, and the mean(MEAN) and max(MAX) classification accuracies on multi-view datasets are employed as the evaluation index.

\subsection{Experiments on textual datasets}
In an attempt to show the superior performance of our framework, the experiments on two multi-view textual datasets (3Source, Cora) are shown in this section. And 1NN classifier is adopted here to classify all testing samples to verify the performances of all DR methods when we have obtained the low-dimensional embedding using all DR methods.

For 3Source dataset, we randomly select 70\% of the samples for each subset as training samples every times. The dimension of embedding obtained by all DR methods all maintains 30 dimensions. We run all DR methods 30 times with different random training samples and testing samples. Table \ref{3Source} shows the MEAN and MAX value on 3Source dataset.
\begin{table}[htbp]
\caption{The classification accuracy on 3Source dataset}
\label{3Source}
\centering  %
\begin{tabular*}{0.45\textwidth}{@{\extracolsep{\fill}}lllll}  
\hline
DR Methods & \multicolumn{2}{c}{Dims=20} & \multicolumn{2}{c}{Dims=30} \\
 & MEAN(\%) & MAX(\%) & MEAN(\%) & MAX(\%)\\
\hline
BLLE & 69.9 & 79.1 & 72.7 & 79.8 \\
BLE & 71.6 & 75.4 & 68.7 & 75.8 \\
CLLE & 77.3 & 88.2 & 78.3 & 85.2 \\
MSE & 79.3 & 90.5 & 79.8 & 91.0 \\
CCA  & 53.8 & 76.4 & 54.7 & 73.5 \\
PMLE & \textbf{83.5} & \textbf{92.1} & \textbf{86.9} & \textbf{92.5}  \\
CMLLE & 82.7 & 90.5 & 81.7 & 91.9 \\
\hline
\end{tabular*}
\end{table}
For Cora dataset, we randomly select 70\% of the samples for each subset as training samples every times. The dimension of embedding obtained by all DR methods all maintains 30 dimensions. We run all DR methods 30 times with different random training samples and testing samples. Table \ref{Cora} shows the MEAN and MAX value on 3Source dataset.
\begin{table}[htbp]
\caption{The classification accuracy on Cora dataset}
\label{Cora}
\centering  %
\begin{tabular*}{0.45\textwidth}{@{\extracolsep{\fill}}lllll}  
\hline
DR Methods & \multicolumn{2}{c}{Dims=20} & \multicolumn{2}{c}{Dims=30} \\
 & MEAN(\%) & MAX(\%) & MEAN(\%) & MAX(\%)\\
\hline
BLLE & 61.3 & 65.6 & 60.9 & 66.7 \\
BLE & 61.7 & 66.3 & 64.7 & 68.5 \\
CLLE & 46.3 & 49.9 & 54.5 & 58.3 \\
MSE & 40.3 & 42.8 & 40.7 & 44.6 \\
CCA  & 71.1 & 73.8 & 71.5 & 74.3 \\
PMLE & 60.6 & 63.6 & 60.0 & 62.3 \\
CMLLE & \textbf{73.8} & \textbf{75.4} & \textbf{74.1} & \textbf{76.8} \\
\hline
\end{tabular*}
\end{table}
Through Tables \ref{3Source}-\ref{Cora}, we can clearly find that our framework outperforms the other DR methods in most situations. Therefore, our framework for multi-view features are more effective. Because our framework can integrate compatible and complementary information from multi-view features, our framework can obtain a more excellent performance.

\subsection{Experiments on images datasets}
In an attempt to show the superior performance of our framework, the experiments on three multi-view images datasets (Yale, ORL, Caltech101) are shown in this section. And 1NN classifier is adopted here to classify all testing samples to verify the performances of all DR methods when we have obtained the low-dimensional embdedding using all DR methods.

\begin{figure*}[htbp]
\centering
\subfigure[Dim=20]{
\centering
\includegraphics[width=0.45\textwidth]{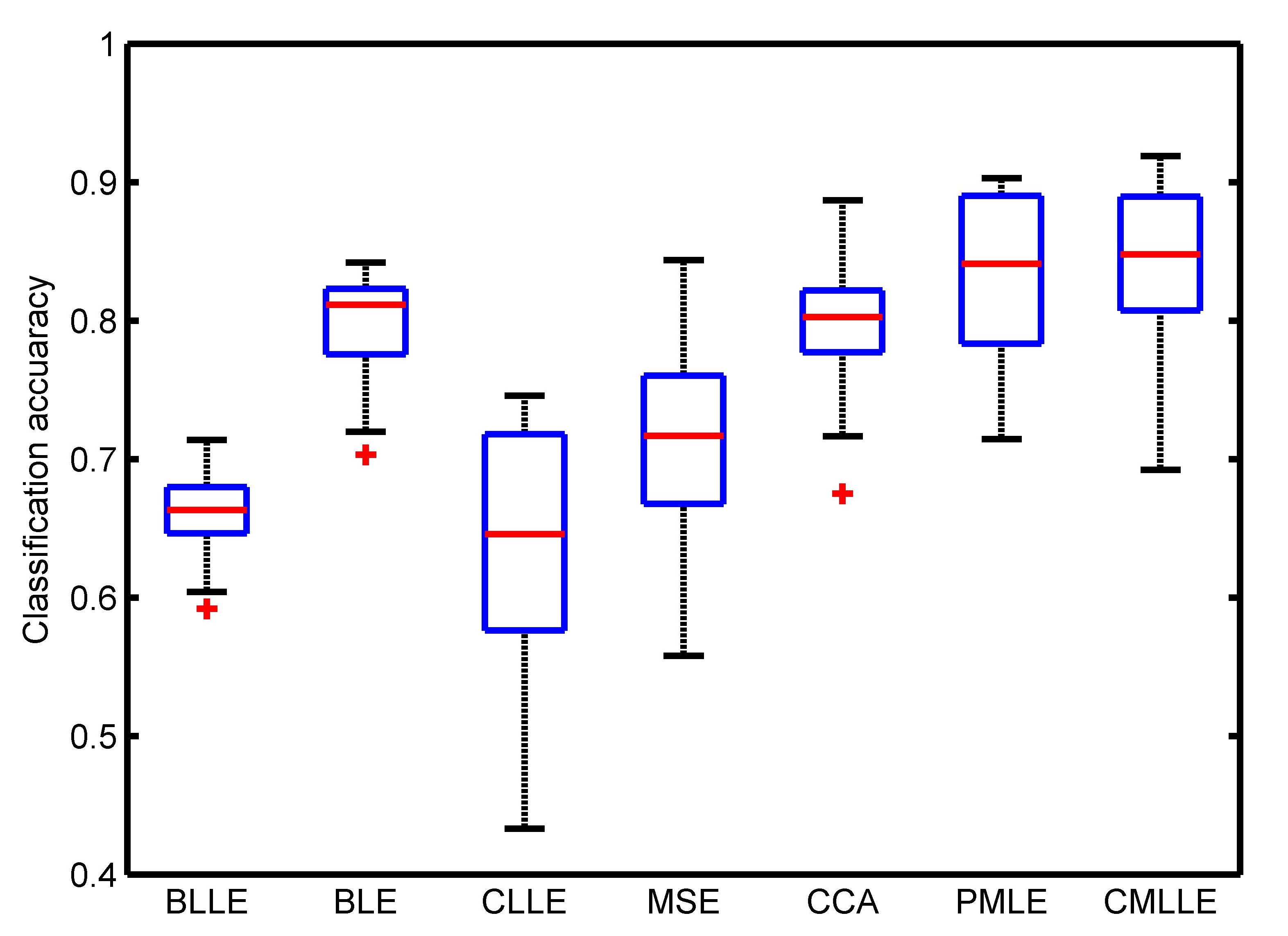}
}%
\subfigure[Dim=30]{
\centering
\includegraphics[width=0.45\textwidth]{yale_dim_20.jpg}
}%
\caption{Classification results on Yale dataset in different dimension}
\label{Yale results}
\end{figure*}

For Yale dataset,we extract gray-scale intensity, local binary patterns and edge direction histogram as 3 views. The dimension of embedding obtained by all DR methods all maintains 20 dimensions and 30 dimensions. We randomly select 80\% of the samples for each subset as training samples every times and run all DR methods 30 times with different random training samples and testing samples. Fig.\ref{Yale results}  shows the accuracy value on Yale dataset.

\begin{figure}[htbp]
\centering
\includegraphics[width=0.45\textwidth]{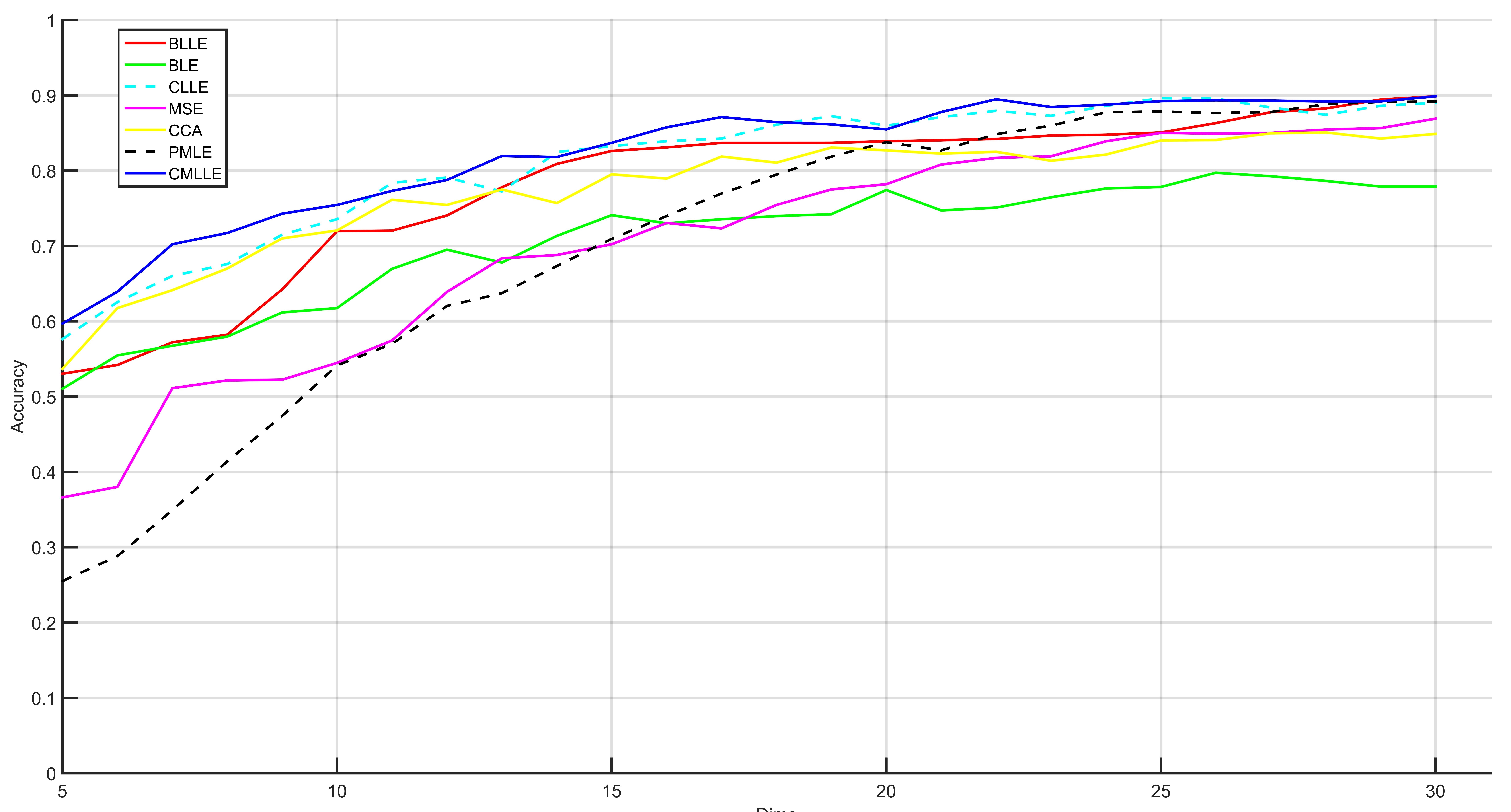}
\caption{The classification accuracy on ORL dataset}
\label{ORL result}
\end{figure}

For ORL dataset, we extract gray-scale intensity, local binary patterns and edge direction histogram as 3 views. The dimension of embedding obtained by all DR methods all maintains from 5 to 30 dimensions. We randomly select 70\% of the samples for each subset as training samples every times and run all DR methods 30 times with different random training samples and testing samples. Fig. \ref{ORL result} shows the mean accuracy values on ORL dataset.

\begin{figure}[htbp]
\centering
\includegraphics[width=0.45\textwidth]{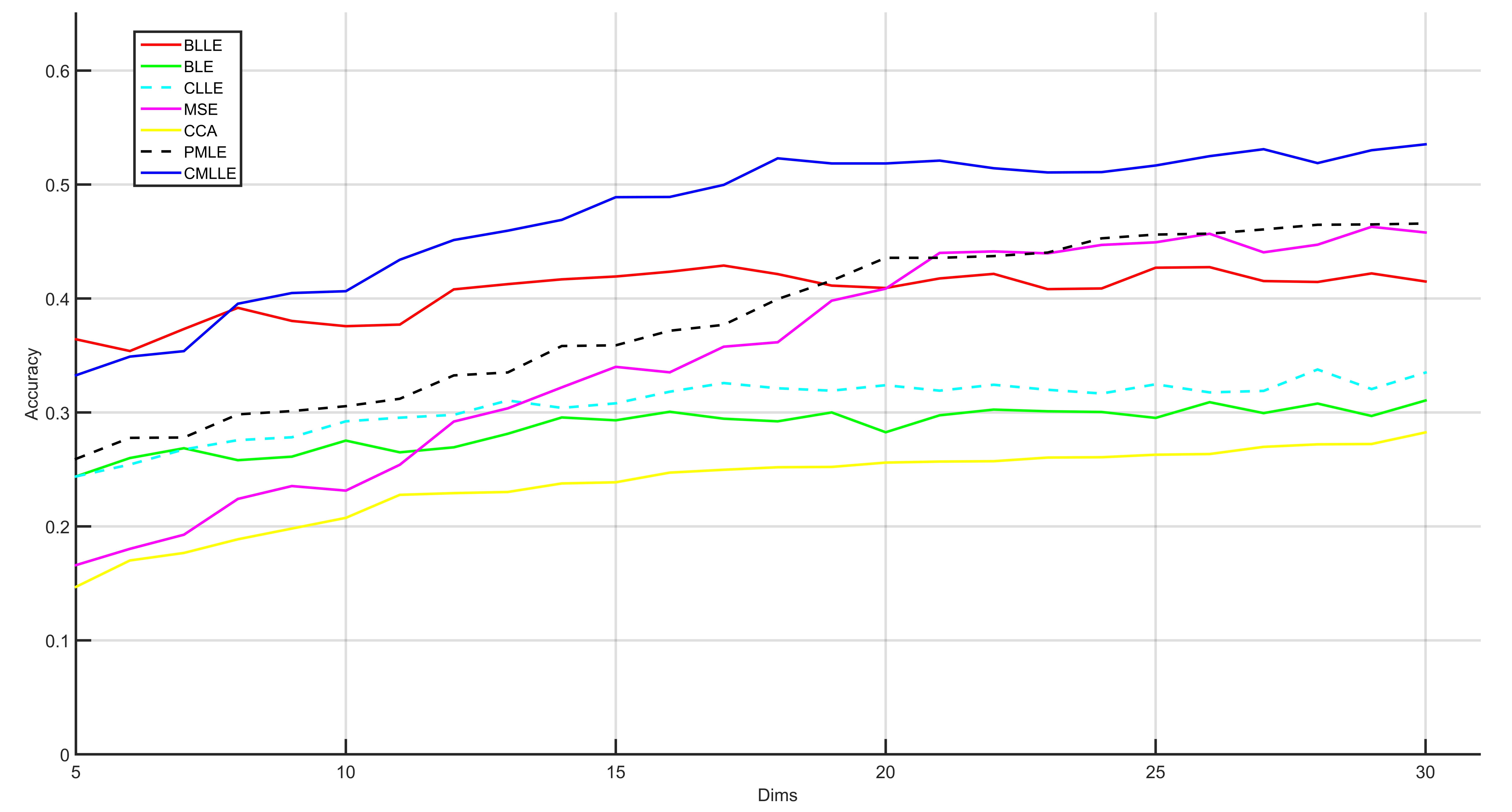}
\caption{The classification accuracy on Caltech101 dataset}
\label{Caltech results}
\end{figure}

For Caltech101 dataset, the first 20 classes are utilized in our experiments. Meanwhile, we extract gist, local binary patterns and edge direction histogram as 3 views. The dimension of embedding obtained by all DR methods maintains from 5 to 30 dimensions. We randomly select 70\% of the samples for each subset as training samples every times and run all DR methods 30 times with different random training samples and testing samples. Fig. \ref{Caltech results} shows the mean accuracy values on Caltech101 dataset.

Through Fig.\ref{Yale results}-\ref{Caltech results}, we can clearly find that our framework outperforms the other DR methods in most situations. And CLLE that concatenating features from different views couldn't gain good performance. Therefore, our framework for multi-view features is more effective. In summary, our framework could integrate compatible and complementary information from multi-view features and obtain more excellent performance.

\subsection{Convergence}
\begin{figure*}[htbp]
\centering
\subfigure[DIM=20 on Cora dataset]{
\centering
\includegraphics[width=0.45\textwidth]{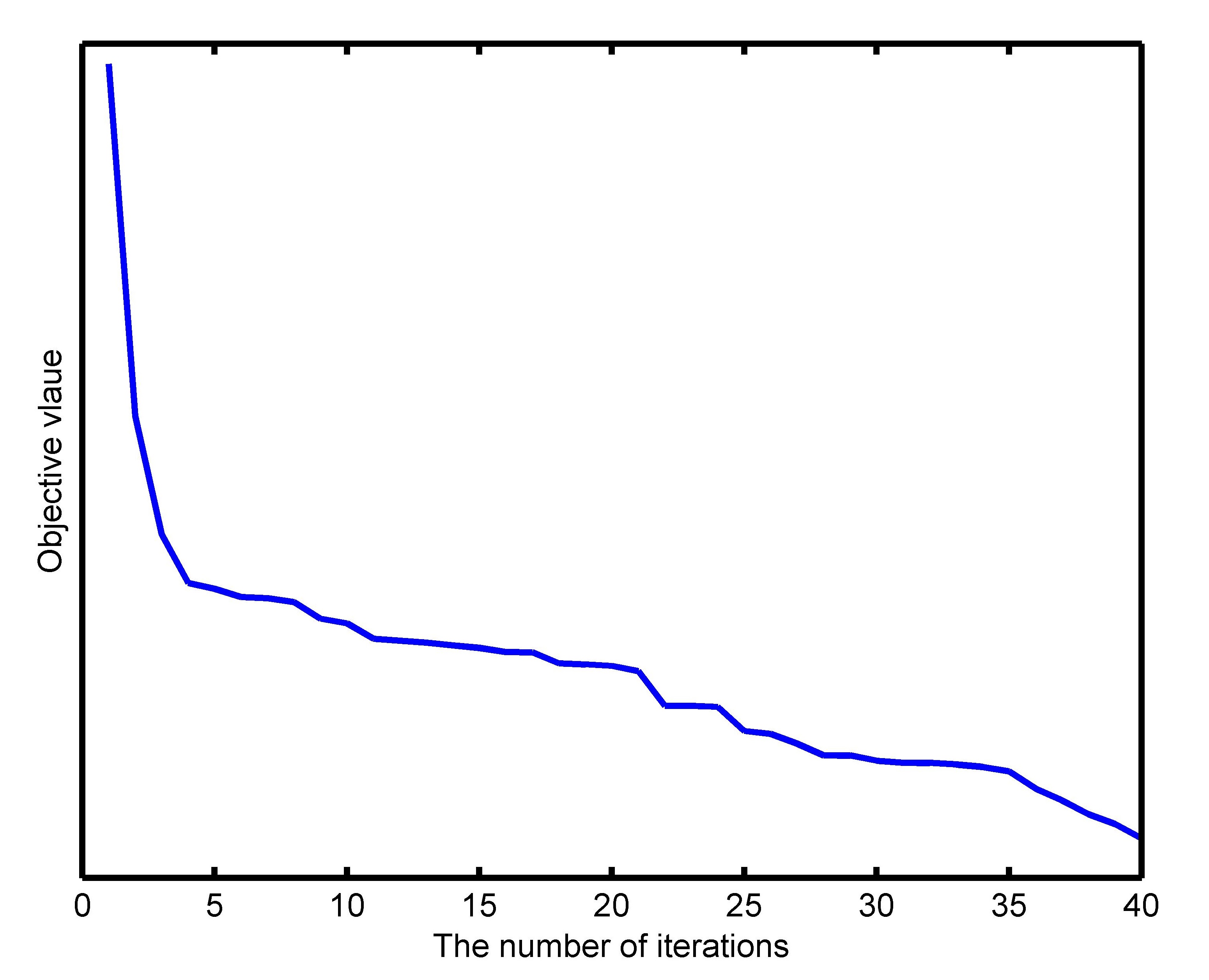}
}
\centering
\subfigure[DIM=30 on Cora dataset]{
\centering
\includegraphics[width=0.45\textwidth]{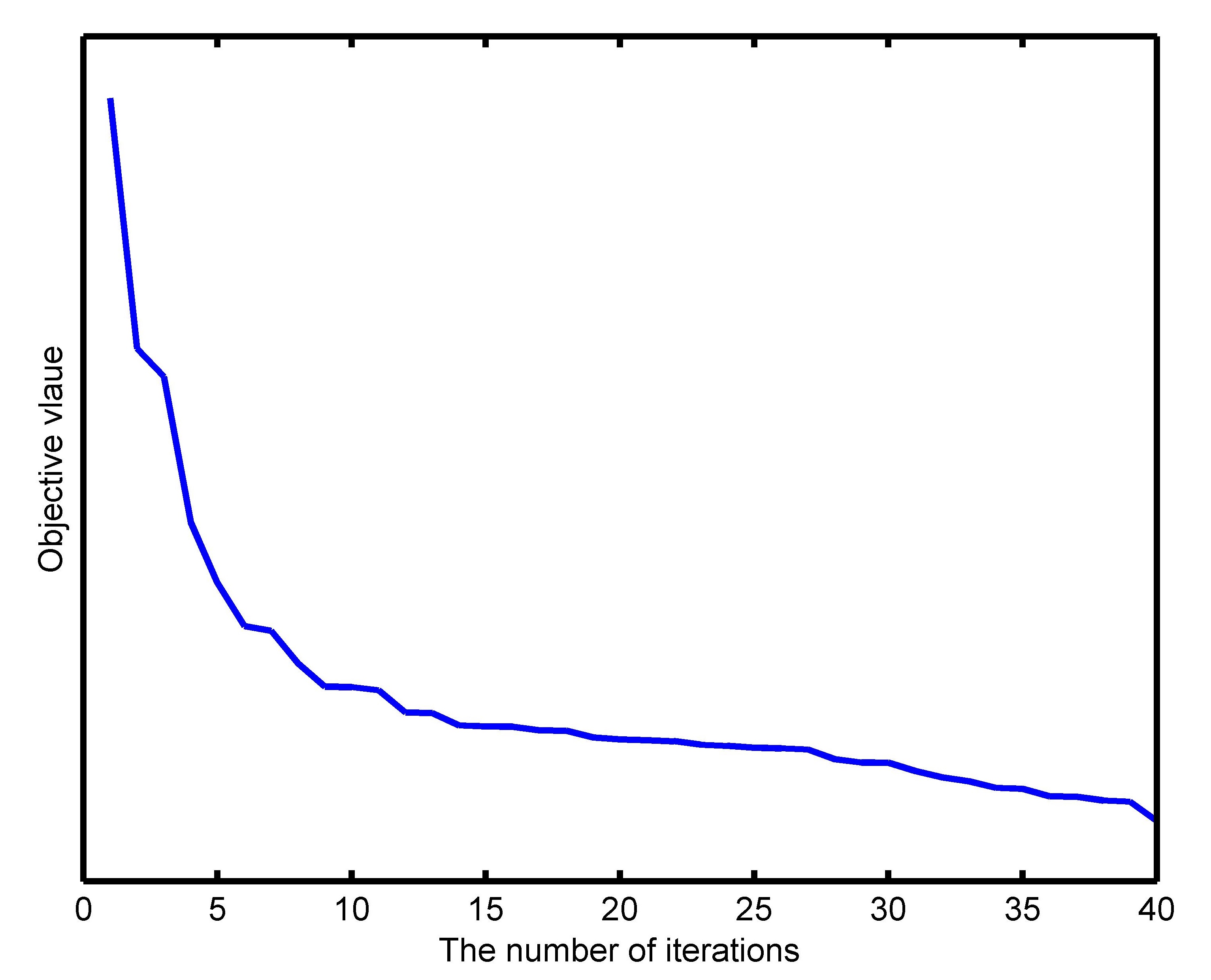}
}

\subfigure[DIM=20 on Caltech101 dataset]{
\centering
\includegraphics[width=0.45\textwidth]{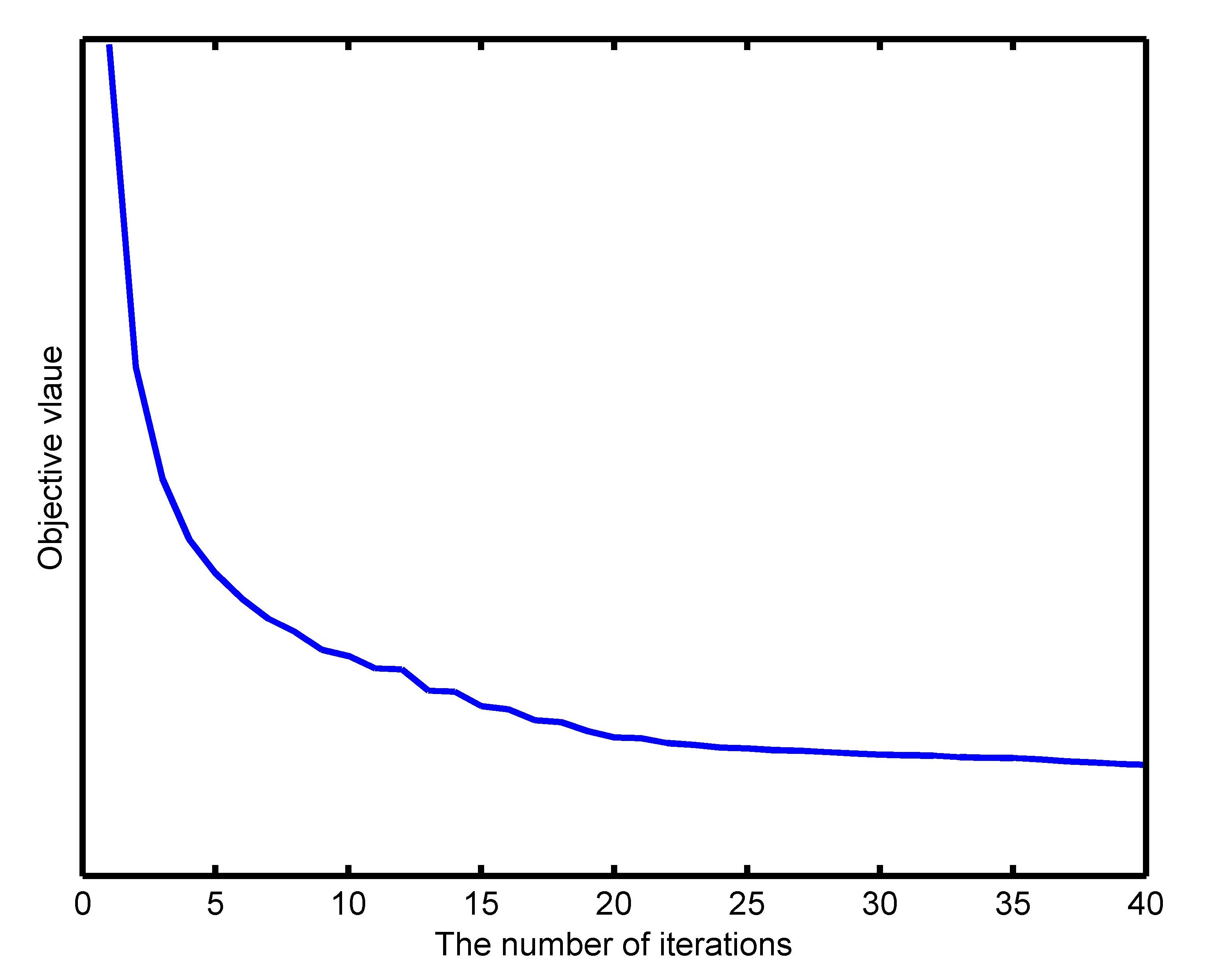}
}
\subfigure[DIM=30 on Caltech101 dataset]{
\centering
\includegraphics[width=0.45\textwidth]{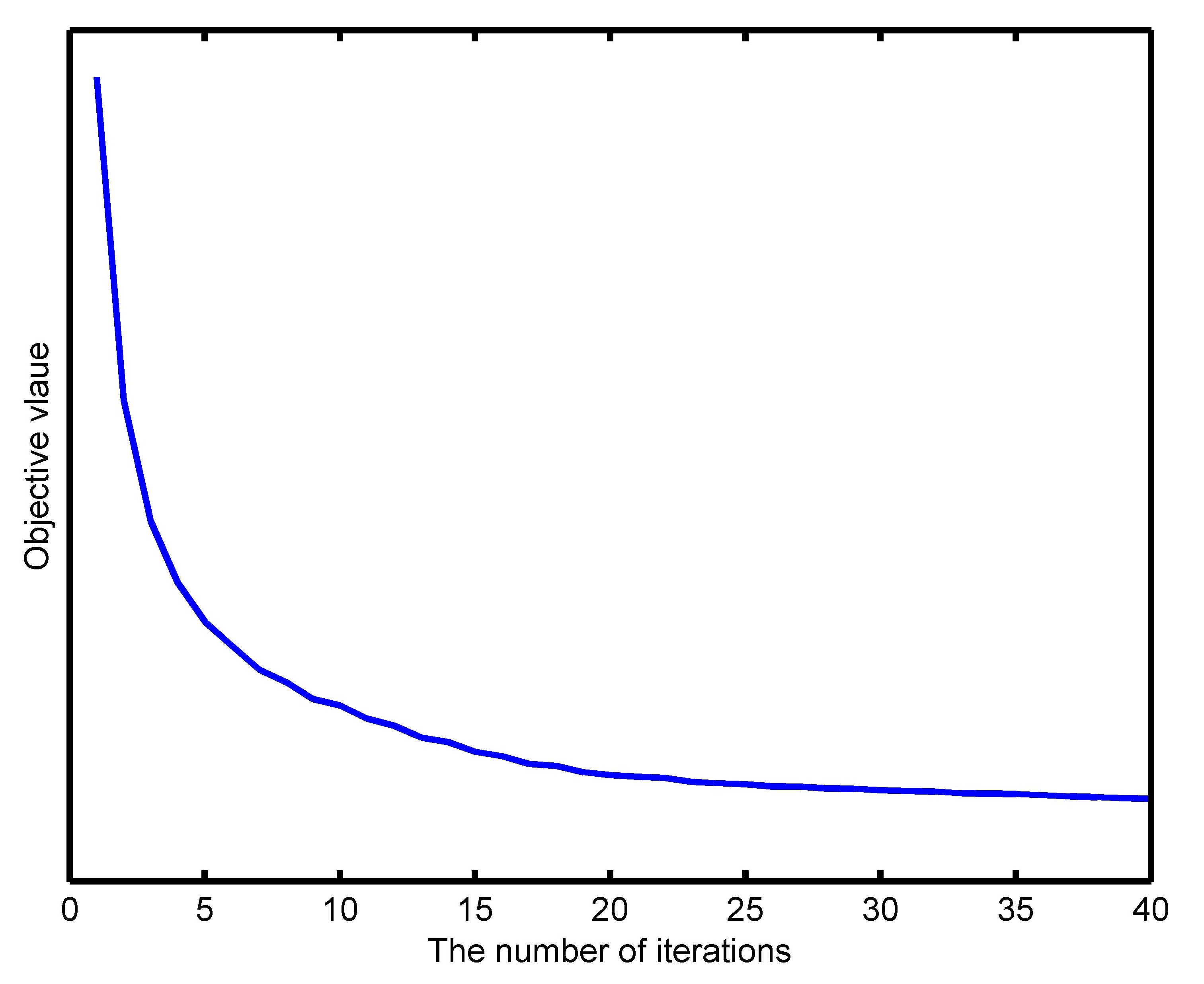}
}

\caption{Objective values of CMLLE on Cora and Caltech101}
\label{convergence}
\end{figure*}

Because our framework adopts an iterative procedure to obtain the optimal solution, it is essential to discuss the convergence in detail. In this section, we summarize the objective values of CMLLE on Cora and Caltech101 datasets according to the above experiments. All the training parameters (such as training numbers, dimensions) can be found above Fig.\ref{convergence}, which summarizes the objective values of Cora and Caltech101 datasets.

We can clearly find in Fig.\ref{convergence} that the curve of the objective values tends to be stable after twenty iterations on the Cora datasets and the objective values tend to be stable after thirty iterations on the Caltech101 datasets. It implies the fact that our framework could converge within a limited number of iterations, and the size of the matrix is an important factor affecting the speed of convergence according to the different iterations numbers of the Cora and Caltech101 datasets.

\section{Conclusion}
In this paper, we propose a novel multi-view learning framework based on similarity consensus, which makes full use of correlations among multi-view features while considering the robustness and scalability of the framework. Besides, it's such a flexible and scalable framework that aims to straightforwardly extend existing single view-based learning methods, including subspace learning methods, kernel methods, and manifold learning methods, into multi-view learning domain by preserving the similarity between different views to capture the low-dimensional embedding for multi-view features. Meanwhile, we provide two schemes based on pairwise and centroid consensus terms in detail and the corresponding algorithms based on iterative alternating strategy are produced to find the optimal solution for our framework. Comprehensive experiments show that our proposed multi-view framework is a comparable and effective multi-view method.

\section*{Acknowledgment}
The authors would like to thank the anonymous reviewers for their insightful comments and the suggestions to significantly improve the quality of this paper. This work was supported by National Natural Science Foundation of PR China(61672130, 61972064) and LiaoNing Revitalization Talents Program(XLYC1806006).

\bibliographystyle{IEEEtran}
\bibliography{IEEEexample}

\begin{IEEEbiography}[{\includegraphics[width=1in,height=1.25in,clip,keepaspectratio]{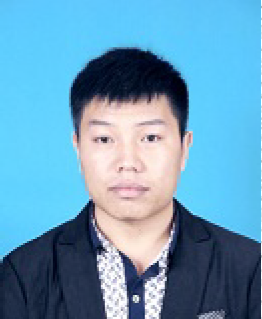}}]{Xiangzhu Meng}
received his BS degree from Anhui University, in 2015. Now he is working towards the PHD degree in School of Computer Science and Technology, Dalian University of Technology, China. His research interests include mulit-view learning, deep learning and computing vision.
\end{IEEEbiography}

\begin{IEEEbiography}[{\includegraphics[width=1in,height=1.25in,clip,keepaspectratio]{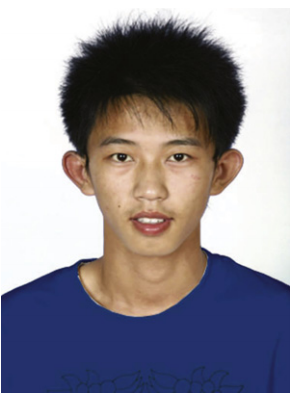}}]{Huibing Wang}
Huibing Wang received the Ph.D. degree in the School of Computer Science and Technology, Dalian University of Technology, Dalian, in 2018. During 2016 and 2017, he is a visiting scholar at the University of Adelaide, Adelaide, Australia. Now, he is a postdoctor in Dalian Maritime University, Dalian, Liaoning, China. He has authored and co-authored more than 20 papers in some famous journals or conferences, including TMM, TITS, TSMCS, ECCV, etc.
Furthermore, he serves as reviewers for TNNLS, Nurocomputing, PR Letters and MTAP, etc. His research interests include computing vision and machine learning
\end{IEEEbiography}

\begin{IEEEbiography}[{\includegraphics[width=1in,height=1.25in,clip,keepaspectratio]{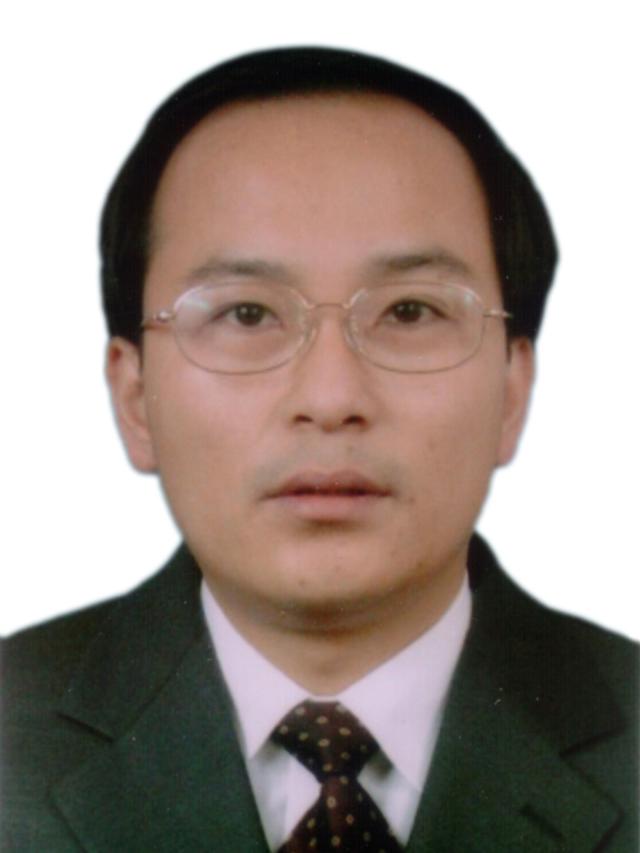}}]{Lin Feng}
received the BS degree in electronic technology from Dalian University of Technology, China, in 1992, the MS degree in power engineering from Dalian University of Technology, China, in 1995,and the PhD degree in mechanical design and theory from Dalian University of Technology, China, in 2004. He is currently a professor and doctoral supervisor in the School of Innovation Experiment, Dalian University of Technology, China. His research interests include intelligent image processing, robotics, data mining, and embedded systems.
\end{IEEEbiography}

\end{document}